\documentclass{article}

\usepackage{microtype}
\usepackage{graphicx}
\usepackage{subfigure}
\usepackage{multirow}
\usepackage{amsmath}
\usepackage{booktabs} 

\usepackage{hyperref}

\usepackage[accepted]{icml2024}

\usepackage{amsmath}
\usepackage{amssymb}
\usepackage{mathtools}
\usepackage{amsthm}

\usepackage[capitalize,noabbrev]{cleveref}

\theoremstyle{plain}

\theoremstyle{definition}

\theoremstyle{remark}

\usepackage[textsize=tiny]{todonotes}

\icmltitlerunning{Learning to Infer Generative Template Programs for Visual Concepts}

\newcommand{\HOLE}{\textit{HOLE}}

\newcommand{\infnets}{$p_{\text{inf}}$}
\newcommand{\gennets}{$p_{\text{gen}}$}
\newcommand{\Target}{\ensuremath{X^*}}
\newcommand{\Objective}{\ensuremath{O}}
\newcommand{\Executor}{\ensuremath{E}}

\newcommand{\VisConcept}{$\widetilde{X}$}
\newcommand{\VisInput}{$x$}

\newcommand{\VisGroup}{$X^G$}
\newcommand{\TemplateGroup}{$TP^G$}
\newcommand{\Template}{\ensuremath{TP}}
\newcommand{\Expansion}{\ensuremath{SE}}
\newcommand{\program}{\ensuremath{z}}
\newcommand{\ProgramGroup}{$Z^G$}

\newcommand{\Metric}{\ensuremath{M}}

\newcommand{\TemplateNet}{$p($\Template$|$\VisGroup$)$}
\newcommand{\ExpansionNet}{$p($\Expansion$|$\Template,\VisInput$)$ }
\newcommand{\ParamNet}{$p($\program$|$\Expansion,\VisInput$)$}

\begin{document}

\twocolumn[
\icmltitle{Learning to Infer Generative Template Programs for Visual Concepts}
  \begin{icmlauthorlist}
   \icmlauthor{R. Kenny Jones}{brown}
   \icmlauthor{Siddhartha Chaudhuri}{adobe}
   \icmlauthor{Daniel Ritchie}{brown}
  \end{icmlauthorlist}

\icmlaffiliation{brown}{Brown University}
\icmlaffiliation{adobe}{Adobe Research}

  \icmlcorrespondingauthor{R. Kenny Jones}{russell\_jones@brown.edu}
  \icmlkeywords{neurosymbolic methods, visual programs, concept learning, program induction, inverse graphics, procedural modelings}

  \vskip 0.3in
]

\printAffiliationsAndNotice{} 

\begin{abstract}

People grasp flexible visual concepts from a few examples.
We explore a neurosymbolic system that learns how to infer programs that capture visual concepts in a domain-general fashion. 
We introduce Template Programs: programmatic expressions from a domain-specific language that specify structural and parametric patterns common to an input concept.
Our framework supports multiple concept-related tasks, including few-shot generation and co-segmentation through parsing.
We develop a learning paradigm that allows us to train networks that infer Template Programs directly from visual datasets that contain concept groupings.
We run experiments across multiple visual domains: 2D layouts, Omniglot characters, and 3D shapes.
We find that our method outperforms task-specific alternatives, and performs competitively against domain-specific approaches for the limited domains where they exist.
\end{abstract}

\section{Introduction}
\label{sec:intro}

Humans understand the visual world through concepts~\cite{murphy2004big}.
Concept-level reasoning allows us to perform a multitude of tasks over a range of situations, even after seeing a new concept only a few times~\cite{likepeople}.
In this paper, we endeavor to endow machines with similar abilities to learn flexible, general purpose visual concepts. 
For instance, to support creative applications, we would like to be able to feed it a small set of visual exemplars and have it synthesize novel generations that match the input concept. 
Or to support analysis tasks, our system should be able to parse the input exemplars into corresponding parts in a consistent fashion.
We desire a system capable of achieving these goals across different visual domains.

Past work has explored systems capable of meeting some of these desiderata~\cite{omniglot}. 
Within methods that approach this problem `symbolically', a common strategy is to induce a structured grammar that explains the input concept set under some criteria (often Bayesian in flavor, e.g. ~\citet{stuhlmuller2010learning}). 
These methods are challenged by the fact that grammar induction is a difficult inverse problem:  these methods often rely on domain specialization or structured inputs, limiting their generality. 
Other attempts have investigated `neural' approaches that learn to perform concept-related tasks~\cite{maml}. 
While these approaches impress in their areas of specialization, such methods can usually only perform a subset of the tasks we are interested in.
They also typically do not generalize well beyond training concepts, unless they have been pretrained on internet scale data.

Working towards domain and task general concept learning, we introduce the \textit{Template Program} framework. 
Our \textit{neurosymbolic} system \textit{learns} how to infer \textit{programs} that capture visual concepts.
Beyond simply parsing concepts, our Template Programs can also be sampled to synthesize new generations conditioned on a group of visual inputs.

Template Programs are structured symbolic objects from a domain-specific language that capture structural and parametric attributes common to a particular concept. 
They admit instantiated programs that accord with these constraints, and convert these programs into visual outputs with a domain-specific executor.
We train networks that learn how to infer Template Programs with a training regime that works across visual domains.
This paradigm requires only a domain-specific language (DSL) and a visual dataset (e.g. images) with concept groupings (e.g. class annotations). 
Our two-step learning approach first pretrains a series of inference networks on synthetic data sampled from the DSL, and then specializes these networks towards the target dataset with a bootstrapped fine-tuning procedure.

We experimentally validate that our method is capable of inferring Template Programs across multiple visual domains: 2D layouts, Omniglot characters, and 3D shapes.
We demonstrate that Template Programs natively support a number of downstream applications, including few-shot generation and co-segmentation.
We are unaware of any other method that is able to perform these tasks in a domain-general fashion, so we compare against either task-specific or domain-specific alternatives.
With respect to task-specific approaches, we find that our neurosymbolic method achieves superior performance.
For the one domain, Omniglot~\cite{bpl}, where task-general methods have been proposed, we compare our domain-general method against domain-specific approaches and find that we are able to achieve competitive performance. 
We release code for our experiments at: https://github.com/rkjones4/TemplatePrograms 

In summary, our contributions are:
\begin{enumerate}
\item \textit{Template Programs}: a neurosymbolic framework for capturing visual concepts with partially specified programs. Our framework supports a range of tasks including few-shot generation and co-segmentation.
\item An unsupervised learning methodology that allows us to train networks that infer Template Programs in a domain-general fashion from visual datasets that contain no annotations beyond concept groupings.
\end{enumerate}

\section{Related Work}
\label{sec:rel_work}

\textbf{Concept Learning  }
Many works have studied concept learning and related tasks~\cite{omniprog}.
A number of methods take a task-specific focus, such as training neural networks to perform few-shot classification~\cite{matchnets,protonets}.
Other approaches have focused instead on generation, under different conditioning frameworks~\cite{rezende2016oneshot,giannone2022fewshot}. 
While these methods perform well in the areas they specialize for, none of them achieve the task-flexibility we desire.

A smaller number of `neural' approaches have investigated how to learn concept representations that support multiple tasks~\cite{neuralstat,hewitt2018variational}.
While these methods often achieve domain-flexibility by learning from visual data directly, they are data-hungry and don't always generalize well to out-of-distribution concepts: we explore this phenomena in our experiments. 
More recently, `foundation' models that train on internet-scale data have shown promise for mitigating these failure modes, but come with other limitations, including massive compute and data requirements~\cite{ruiz2022dreambooth}.

Most relevant to our approach are methods that learn structured task-general representations.
To the best of our knowledge, such methods have only been successfully developed for stroke-based drawing domains.
~\citet{bpl} (\textit{BPL}) fit a structured hierarchical model of handwritten character production to human stroke data under a Bayesian framing, achieving human-level performance across generative and discriminative tasks. 
~\citet{gns} (\textit{GNS}) extend this framework with a neurosymbolic method, where the distribution and correlation of strokes are modeled with learned networks.
While these approaches demonstrate impressive performance, their design is specialized for datasets such as Omniglot; we are unaware of any successful attempts to generalize these approaches to other domains. 
Of note, DooD is a related neurosymbolic approach that is similarly specialized for drawing domains, but has shown the capability to generalize across drawing-related datasets~\cite{dood}.
As our proposed framework is designed to maintain domain-generality, outperforming these specialists is not our goal. 
That said, we experimentally compare our approach against BPL and GNS on Omniglot and find that we are able to largely match their performance for few-shot generation and co-segmentation tasks.

\textbf{Inverse Procedural Modeling  }
While not often presented as concept learners, methods within computer graphics have made progress on related problems under the framing of inverse procedural modeling. 
These methods aim to infer structured symbolic objects that explain visual inputs, e.g. procedural models that produce visual outputs when executed.
A typical framing these approaches take is to induce a grammar with a bottom-up procedure, e.g. Bayesian merging~\cite{BayesianProgramMerging}. 
These techniques have demonstrated success across many visual domains, including plants~\cite{InverseLSystems,guo2020inverse} and buildings~\cite{nishida_proc, Martinovic_prog, NBG18, InverseProceduralArchitecture}, and some can even induce more general probabilistic programs~\cite{ProcmodLearn}.
Unfortunately, these methods lack the generality we desire: they are not able to induce grammars outside of their specialized domains and often require structured input data. 
In contrast, our approach uses learned networks that guide a top-down inference procedure.

Relatedly, another line of work has explored how to infer procedural models that explain a single visual input; we refer to this task as visual program induction (VPI)~\cite{NcsgSTAR}.  
Though some VPI methods are non-learning based, relying on search and heuristics~\cite{du2018inversecsg,zoneGraphs}, recent methods have investigated learning-based solutions for this task.
While many of these VPI approaches are specialized for domains of interest such as 3D CAD modeling~\cite{xu2022skexgen,tian2019learning}, others have proposed learning techniques that work over multiple domains~\cite{jones2022PLAD,ellis2019write,spiral,MemoizedWakeSleep,ganeshan2023coref}.
Similar to these approaches, we aim to learn networks that can infer programmatic objects across domains, but instead of inferring a \textit{single} program that explains one input, we aim to infer a Template Program that captures a \textit{group} of visual inputs.

A few works have considered how partial program constructs can aid in program synthesis.
Sketch uses partial programs to constrain a SAT-based inductive synthesis procedure~\cite{solar2008program}.
While this framing is compelling for discrete domains, adapting this technique for more complex visual domains often requires first solving a primitive decomposition problem~\cite{handdrawn}.
DreamCoder is a system that takes as input a corpus of tasks and a base DSL, and from this uses an iterative bootstrapping procedure to find a library of abstraction functions~\cite{DreamCoder}.
While these discovered abstractions may contain holes, they cannot be used to perform  concept-related tasks.
In contrast to both of these works, which use partial program intermediates to find a deterministic program that matches a single input,
our method aims to find a partial programs that explains a collection of visual inputs.

\begin{figure*}[t!]
\centering
  \includegraphics[width=\textwidth]{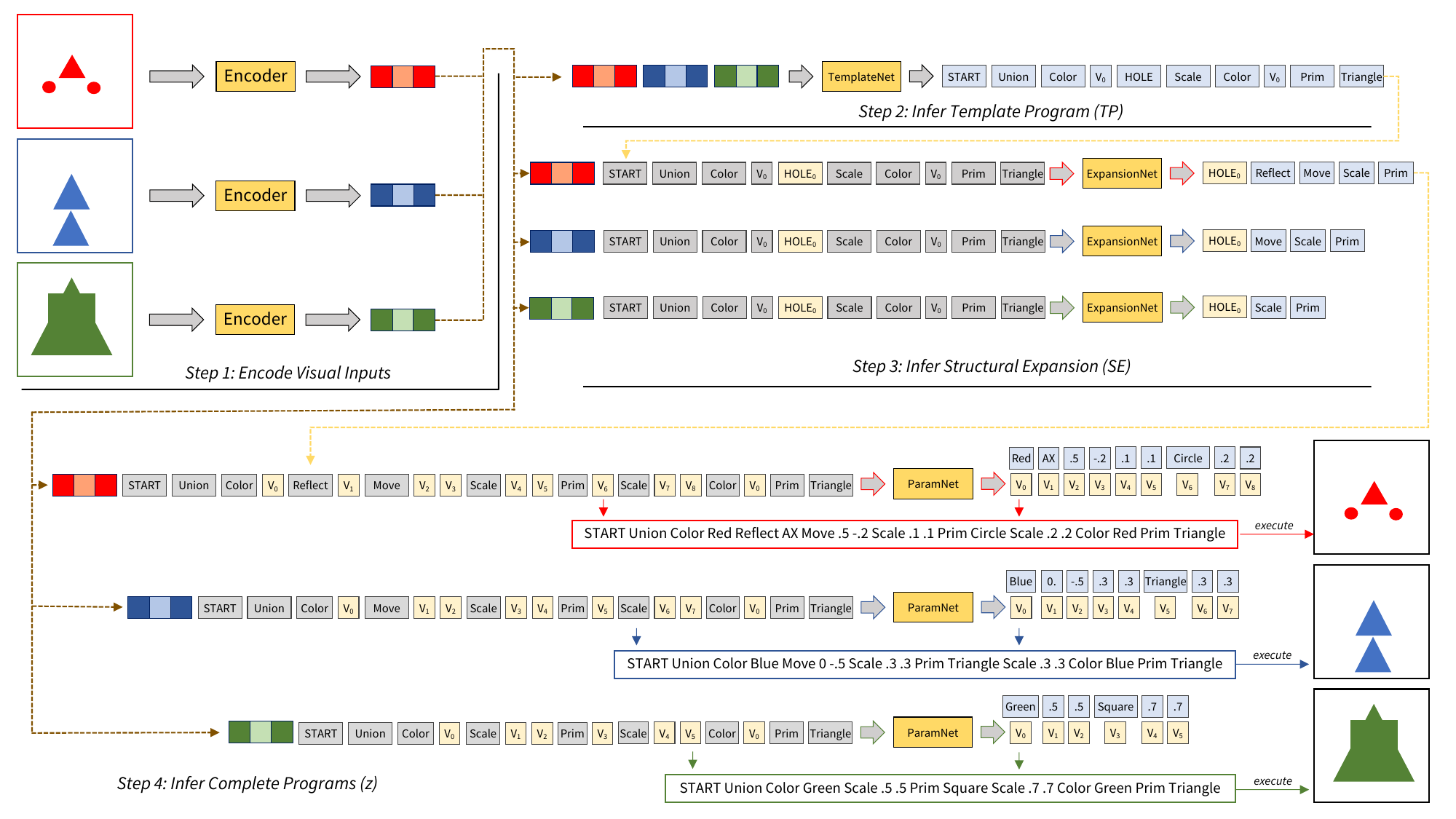}
  \vspace{-1em}
  \caption{Our inference process. First, a group of visual inputs are encoded (Step 1). Next, our TemplateNet uses these encodings to infer a Template Program (\Template, Step 2). The~\Template~and each encoding are then sent to the ExpansionNet to produce a Structural Expansion ~(\Expansion) for each input (Step 3), which are finally passed to the ParamNet to produce a set of complete programs that explain the inputs (Step 4).} 
  \label{fig:method}
\end{figure*}

\section{Method}
\label{sec:method}

Our framework learns how to infer Template Programs (Section~\ref{sec:met_template}) that capture visual concepts. 
We describe our inference networks in Section~\ref{sec:met_inf} and our learning paradigm in Section~\ref{sec:met_learn}. 

\subsection{Template Programs}
\label{sec:met_template}

Given a collection of related visual inputs, our goal is to find a symbolic structure capable of representing this group as a concept.
This structure must be able to account for both (i) the shared attributes across the group and (ii) the allowable divergences that differentiate various group members.

Towards this goal, we introduce \textit{Template Programs} to represent visual concepts. 
A Template Program~(\Template) is a partial program specification from a domain-specific language (DSL).
We assume this DSL is a functional language, where each function takes other functions or parameter arguments as input.
Template Programs admit fully instantiated \textit{programs} (\program).
These programs can be run through a domain-specific executor (\Executor) to produce visual outputs.

Template Programs are composed of a hierarchy of function calls (i.e an expression tree) and are optionally allowed to define relationships between parameter arguments (e.g. variable reuse).
All instantiations from a Template Program must invoke the specified functions and use the described relations.
To allow instantiations to vary structurally (i.e. use different functions), we introduce a special~\HOLE~construct.
Each~\HOLE~in the Template Program can be filled in with an arbitrary expression tree.
This process creates a \textit{Structural Expansion} (\Expansion), which completely specifies the function call sequence of an instantiation. 
Any function parameters that lack a specified relation in the~\Expansion~are allowed to differ freely in the output programs.

\subsection{Inference Networks}
\label{sec:met_inf}

We use a learning-based approach to infer Template Programs and their instantiations.
Given a group of visual inputs~\VisGroup~from some concept~\VisConcept, our goal is to infer a Template Program~\Template, such that for each~\VisInput~in~\VisGroup~there is a program instantiation~\program~from~\Template~so that~\Executor(\program)~$=$~\VisInput.

We solve this difficult inverse structured prediction problem with a series of inference networks~\infnets~that we depict in Figure~\ref{fig:method}.
To start, each~\VisInput~is converted into a latent code with a domain-specific visual encoder (e.g. a 2D CNN for image inputs).
These latent codes are then passed through a series of auto-regressive networks, explained below. 

The TemplateNet,~\TemplateNet, is responsible for inferring Template Programs. 
Attending over all of the latent codes from~\VisGroup~as conditioning information, it autoregressively predicts a series of tokens that form the Template Program.
We linearize this composition of functions with prefix notation.
Using Figure~\ref{fig:method} as reference, these tokens are either (i)~functions from the DSL (\texttt{SCALE}), (ii)~\HOLE~tokens, or (iii)~parametric relations, such as static variable assignment~(\texttt{Triangle}) or variable reuse~(\texttt{$V_0$}).

Given the inferred~\Template, we use the ExpansionNet and ParamNet to instantiate a complete program~\program.
The ExpansionNet,~\ExpansionNet, conditions on~\Template~along with a single visual input~\VisInput, and autoregressively produces a~\Expansion~by filling in~\HOLE~tokens with a series of functions.
This~\Expansion~is then reformatted to expose any free parameters and their relations.
The ParamNet,~\ParamNet, conditions on this representation and the same visual input~\VisInput~in order to autoregressively predict the value of each parameter which instantiates a complete program~\program.

\subsection{Learning Paradigm}
\label{sec:met_learn}
How can we train our inference networks?
With ground-truth program annotations, we could employ supervised learning, but  datasets with this level of annotation do not exist. 
As our goal is to design a domain-general framework, our problem formulation assumes the following as input: a target dataset of interest \Target~and a relevant DSL.
We assume that we can sample groups of visual concepts from this dataset (e.g. by using class annotations), but otherwise assume the visual data is unstructured.
Under these assumptions, we employ a two-step process: we first initialize our networks by pretraining on synthetic data sampled from the DSL, and then we  specialize~\infnets~towards~\Target~with bootstrapped finetuning.

\textbf{Synthetic Pretraining  }
We implement each autoregressive network within~\infnets~as a Transformer decoder with causal masking (where the conditioning information varies across networks).
With paired (input, output) data, each of these networks can be trained with maximum likelihood updates (i.e. cross-entropy loss).
We can produce (input, output) pairs for all of our networks if we have an associated (\VisGroup, \TemplateGroup, \ProgramGroup) group, where targets for the ExpansionNet and ParamNet can be derived by comparing the \TemplateGroup~to each \program~$\in$~\ProgramGroup~(further details in Appendix~\ref{sec:app_teacher_force}).

One way to produce paired data is to generate it synthetically. 
Following previous VPI approaches~\cite{tian2019learning,sharma2018csgnet}, we sample synthetic data from our DSL and use it to pretrain our inference networks in a supervised setting. 
At a high level, this sampling procedure invokes the following steps: (1) sample a full program from the DSL (e.g. by stochastically expanding the grammar), (2) convert the full program into a~\TemplateGroup~(e.g. by collapsing random expression trees into~\HOLE~tokens and randomly assigning parameter relations), (3) sampling a group of programs~\ProgramGroup~from the~\TemplateGroup~(e.g. through random expansion) and recording their executions,~\VisGroup~=~(\Executor(\program)~$\forall$~\program~$\in$~\ProgramGroup).

\textbf{Bootstrapped Finetuning  }
While synthetic pretraining attunes~\infnets~to the DSL, it produces overly general networks that make inaccurate predictions when run over concepts from~\Target.
To specialize~\infnets~towards~\Target, we develop an unsupervised bootstrapped finetuning approach that generalizes the PLAD framework designed for single-input, deterministic programs~\cite{jones2022PLAD}.

Our algorithm oscillates between inference and training steps.
In each inference step, we run~\infnets~over groups of visual inputs~\VisGroup~drawn from concepts in the target dataset~\VisConcept~$\in$~\Target. 
We run a beam-search to find the Template Program whose instantiations best match~\VisGroup~under an objective~\Objective~(Eq.~\ref{eq:obj}).
For each~\VisGroup, we record the best inferred (\TemplateGroup,~\ProgramGroup) pair for use in the training step.

The training step uses this paired data to finetune~\infnets. 
Specifically, we convert (\VisGroup,~\TemplateGroup,~\ProgramGroup) inferred groups into paired training data for~\infnets~under different self-supervised learning formulations.
In the self-training (\textit{ST}) formulation, we leave the group as is.
In the latent execution self-training formulation (\textit{LEST}), we replace~\VisGroup~by executing each program in~\ProgramGroup.
Our wake-sleep formulation (\textit{WS}) first trains a generative model~\gennets~(Appendix~\ref{sec:app_gcond_net}).
This model is a modified variant of~\infnets, where the visual latent codes are masked out, so that visual information does not affect the conditioning.
We train~\gennets~to model the inferred (~\TemplateGroup,~\ProgramGroup) data, and then we sample a collection of synthesized (\TemplateGroup,~\ProgramGroup) pairings from the network. 
Finally, we produce an associated~\VisGroup~for each generation by employing our program executor, following the same procedure as in LEST. 

From these three self-supervised approaches (ST, LEST, WS), we get three distinct datasets of (\VisGroup, \TemplateGroup, \ProgramGroup) groups. 
We use these datasets to finetune~\infnets, using the same maximum likelihood updates as in our synthetic pretraining phase.
We randomly sample batches from each of these datasets in a training loop until we reach convergence with respect to concepts from the validation set of~\Target.

\begin{figure*}[t!]
    \centering
    \tiny
    \setlength{\tabcolsep}{.5pt}
    \begin{tabular}{cccccccccccccccccc}
          \multicolumn{18}{c}{\textit{2D Primitive Layouts}} \\
        \raisebox{1.0em}{Inp} \hspace{.25em} &
        \includegraphics[{width=.059\linewidth}]{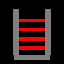} &
        \includegraphics[{width=.059\linewidth}]{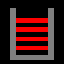} &
        \includegraphics[{width=.059\linewidth}]{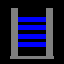} &
        \includegraphics[{width=.059\linewidth}]{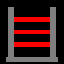} &
        \includegraphics[{width=.059\linewidth}]{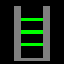} &
        \hspace{.75em}
        &
        \includegraphics[{width=.059\linewidth}]{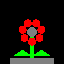} &
        \includegraphics[{width=.059\linewidth}]{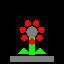} &
        \includegraphics[{width=.059\linewidth}]{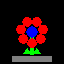} &
        \includegraphics[{width=.059\linewidth}]{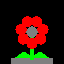} &
        \includegraphics[{width=.059\linewidth}]{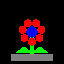} &
         \hspace{.75em}
        &
        \includegraphics[{width=.059\linewidth}]{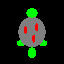} &
        \includegraphics[{width=.059\linewidth}]{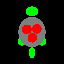} &
        \includegraphics[{width=.059\linewidth}]{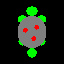} &
        \includegraphics[{width=.059\linewidth}]{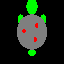} &
        \includegraphics[{width=.059\linewidth}]{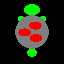} \\

        \raisebox{1.0em}{Seg} \hspace{.75em} &
        \includegraphics[{width=.059\linewidth}]{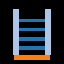} &
        \includegraphics[{width=.059\linewidth}]{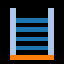} &
        \includegraphics[{width=.059\linewidth}]{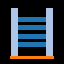} &
        \includegraphics[{width=.059\linewidth}]{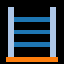} &
        \includegraphics[{width=.059\linewidth}]{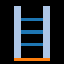} &
        \hspace{.75em}
        &
        \includegraphics[{width=.059\linewidth}]{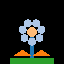} &
        \includegraphics[{width=.059\linewidth}]{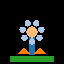} &
        \includegraphics[{width=.059\linewidth}]{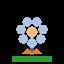} &
        \includegraphics[{width=.059\linewidth}]{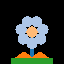} &
        \includegraphics[{width=.059\linewidth}]{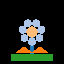} &
         \hspace{.75em}
        &
        \includegraphics[{width=.059\linewidth}]{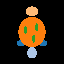} &
        \includegraphics[{width=.059\linewidth}]{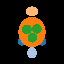} &
        \includegraphics[{width=.059\linewidth}]{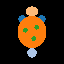} &
        \includegraphics[{width=.059\linewidth}]{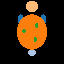} &
        \includegraphics[{width=.059\linewidth}]{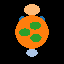} \\
        \raisebox{1.0em}{Gen} \hspace{.75em} &
        \includegraphics[{width=.059\linewidth}]{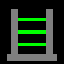} &
        \includegraphics[{width=.059\linewidth}]{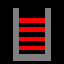} &
        \includegraphics[{width=.059\linewidth}]{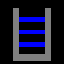} &
        \includegraphics[{width=.059\linewidth}]{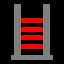} &
        \includegraphics[{width=.059\linewidth}]{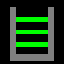} &
        \hspace{.75em}
        &
        \includegraphics[{width=.059\linewidth}]{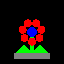} &
        \includegraphics[{width=.059\linewidth}]{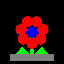} &
        \includegraphics[{width=.059\linewidth}]{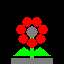} &
        \includegraphics[{width=.059\linewidth}]{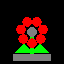} &
        \includegraphics[{width=.059\linewidth}]{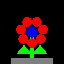} &
         \hspace{.75em}
        &
        \includegraphics[{width=.059\linewidth}]{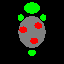} &
        \includegraphics[{width=.059\linewidth}]{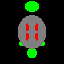} &
        \includegraphics[{width=.059\linewidth}]{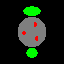} &
        \includegraphics[{width=.059\linewidth}]{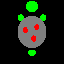} &
        \includegraphics[{width=.059\linewidth}]{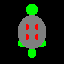} \\
        \multicolumn{18}{c}{\textit{Omniglot Characters}} \\
         \raisebox{1.0em}{Inp} \hspace{.25em} &
        \includegraphics[{width=.059\linewidth}]{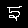} &
        \includegraphics[{width=.059\linewidth}]{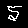} &
        \includegraphics[{width=.059\linewidth}]{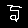} &
        \includegraphics[{width=.059\linewidth}]{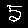} &
        \includegraphics[{width=.059\linewidth}]{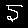} &
        \hspace{.75em}
        &
        \includegraphics[{width=.059\linewidth}]{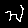} &
        \includegraphics[{width=.059\linewidth}]{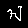} &
        \includegraphics[{width=.059\linewidth}]{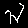} &
        \includegraphics[{width=.059\linewidth}]{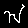} &
        \includegraphics[{width=.059\linewidth}]{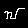} &
         \hspace{.75em}
        &
        \includegraphics[{width=.059\linewidth}]{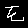} &
        \includegraphics[{width=.059\linewidth}]{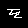} &
        \includegraphics[{width=.059\linewidth}]{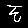} &
        \includegraphics[{width=.059\linewidth}]{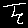} &
        \includegraphics[{width=.059\linewidth}]{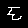} \\

        \raisebox{1.0em}{Seg} \hspace{.75em} &
        \includegraphics[{width=.059\linewidth}]{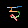} &
        \includegraphics[{width=.059\linewidth}]{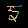} &
        \includegraphics[{width=.059\linewidth}]{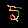} &
        \includegraphics[{width=.059\linewidth}]{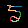} &
        \includegraphics[{width=.059\linewidth}]{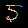} &

        \hspace{.75em}
        &
        \includegraphics[{width=.059\linewidth}]{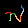} &
        \includegraphics[{width=.059\linewidth}]{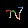} &
        \includegraphics[{width=.059\linewidth}]{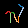} &
        \includegraphics[{width=.059\linewidth}]{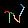} &
        \includegraphics[{width=.059\linewidth}]{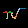} &
         \hspace{.75em}
        &
        \includegraphics[{width=.059\linewidth}]{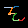} &
        \includegraphics[{width=.059\linewidth}]{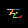} &
        \includegraphics[{width=.059\linewidth}]{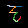} &
        \includegraphics[{width=.059\linewidth}]{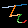} &
        \includegraphics[{width=.059\linewidth}]{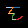}  \\
        \raisebox{1.0em}{Gen} \hspace{.75em} &
        \includegraphics[{width=.059\linewidth}]{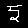} &
        \includegraphics[{width=.059\linewidth}]{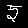} &
        \includegraphics[{width=.059\linewidth}]{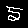} &
        \includegraphics[{width=.059\linewidth}]{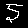} &
        \includegraphics[{width=.059\linewidth}]{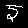} &
        \hspace{.75em}
        &
        \includegraphics[{width=.059\linewidth}]{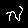} &
        \includegraphics[{width=.059\linewidth}]{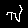} &
        \includegraphics[{width=.059\linewidth}]{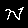} &
        \includegraphics[{width=.059\linewidth}]{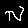} &
        \includegraphics[{width=.059\linewidth}]{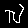} &
         \hspace{.75em}
        &
        \includegraphics[{width=.059\linewidth}]{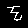} &
        \includegraphics[{width=.059\linewidth}]{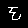} &
        \includegraphics[{width=.059\linewidth}]{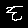} &
        \includegraphics[{width=.059\linewidth}]{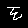} &
        \includegraphics[{width=.059\linewidth}]{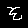} \\
         \multicolumn{18}{c}{\textit{3D Shape Structures}} \\
         \raisebox{1.0em}{Inp} \hspace{.25em} &
        \includegraphics[{width=.059\linewidth}]{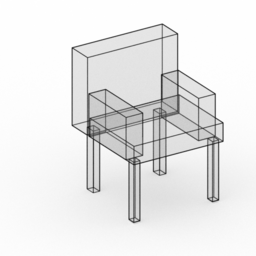} &
        \includegraphics[{width=.059\linewidth}]{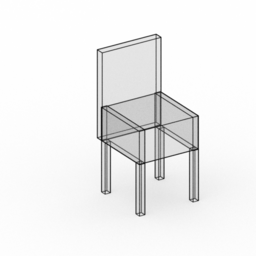} &
        \includegraphics[{width=.059\linewidth}]{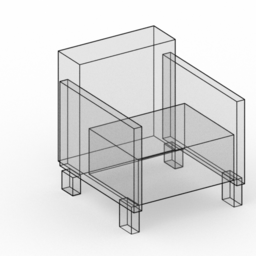} &
        \includegraphics[{width=.059\linewidth}]{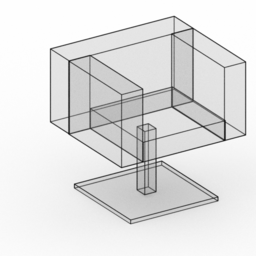} &
        \includegraphics[{width=.059\linewidth}]{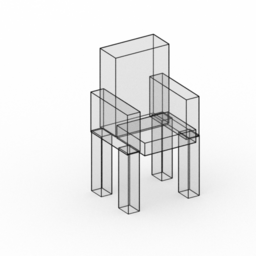} &
        \hspace{.75em}
        &
        \includegraphics[{width=.059\linewidth}]{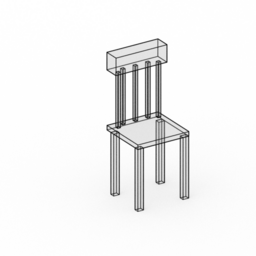} &
        \includegraphics[{width=.059\linewidth}]{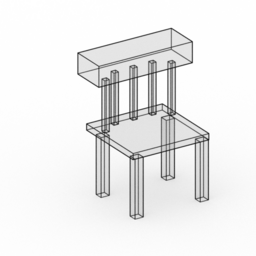} &
        \includegraphics[{width=.059\linewidth}]{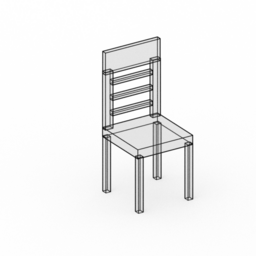} &
        \includegraphics[{width=.059\linewidth}]{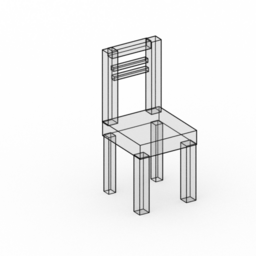} &
        \includegraphics[{width=.059\linewidth}]{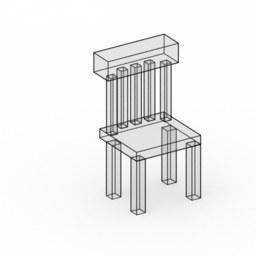} &
         \hspace{.75em}
        &
        \includegraphics[{width=.059\linewidth}]{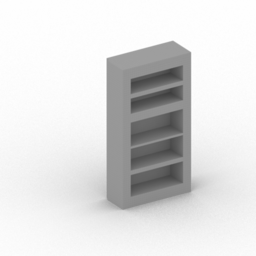} &
        \includegraphics[{width=.059\linewidth}]{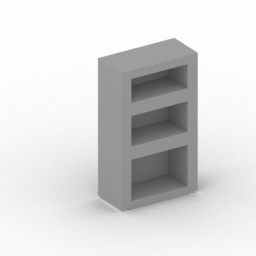} &
        \includegraphics[{width=.059\linewidth}]{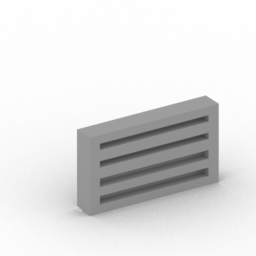} &
        \includegraphics[{width=.059\linewidth}]{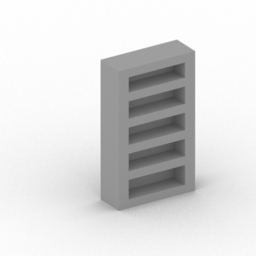} &
        \includegraphics[{width=.059\linewidth}]{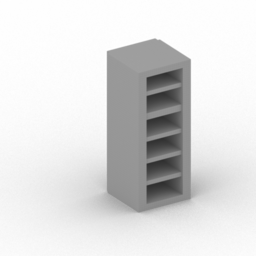} \\

        \raisebox{1.0em}{Seg} \hspace{.75em} &
        \includegraphics[{width=.059\linewidth}]{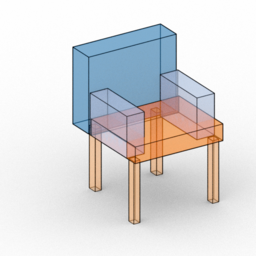} &
        \includegraphics[{width=.059\linewidth}]{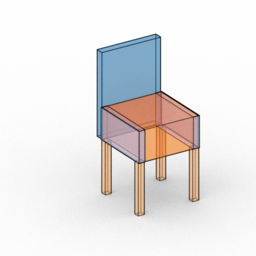} &
        \includegraphics[{width=.059\linewidth}]{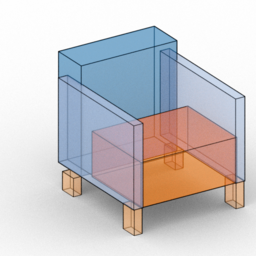} &
        \includegraphics[{width=.059\linewidth}]{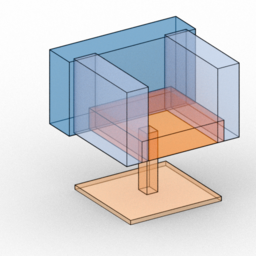} &
        \includegraphics[{width=.059\linewidth}]{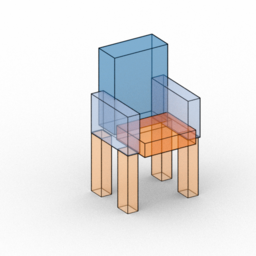} &
        \hspace{.75em}
        &
        \includegraphics[{width=.059\linewidth}]{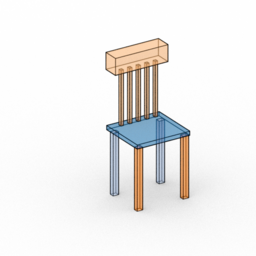} &
        \includegraphics[{width=.059\linewidth}]{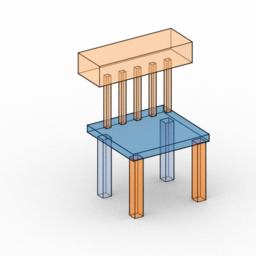} &
        \includegraphics[{width=.059\linewidth}]{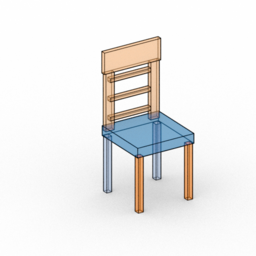} &
        \includegraphics[{width=.059\linewidth}]{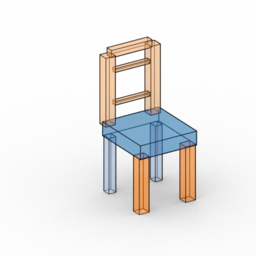} &
        \includegraphics[{width=.059\linewidth}]{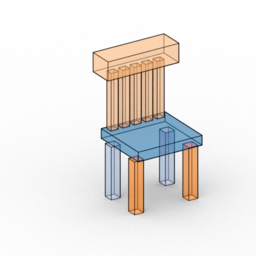} &
         \hspace{.75em}
        &
        \includegraphics[{width=.059\linewidth}]{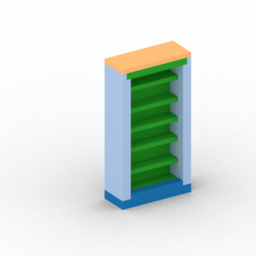} &
        \includegraphics[{width=.059\linewidth}]{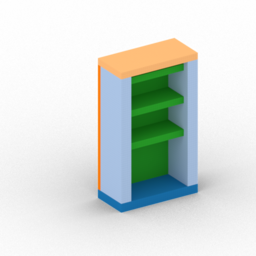} &
        \includegraphics[{width=.059\linewidth}]{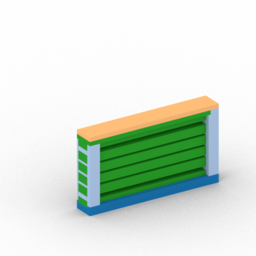} &
        \includegraphics[{width=.059\linewidth}]{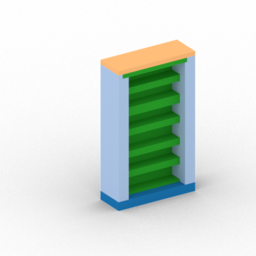} &
        \includegraphics[{width=.059\linewidth}]{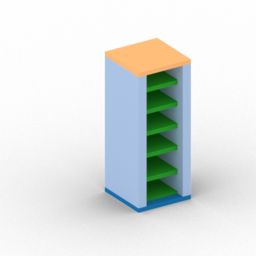} \\
        \raisebox{1.0em}{Gen} \hspace{.75em} &
        \includegraphics[{width=.059\linewidth}]{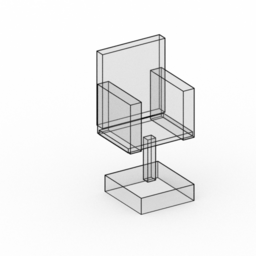} &
        \includegraphics[{width=.059\linewidth}]{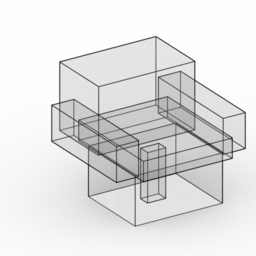} &
        \includegraphics[{width=.059\linewidth}]{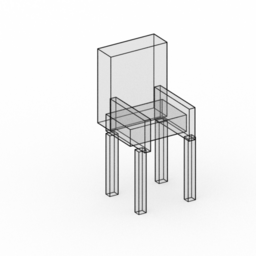} &
        \includegraphics[{width=.059\linewidth}]{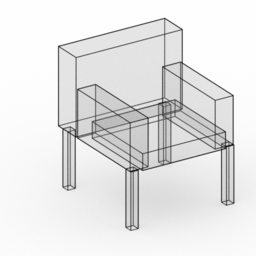} &
        \includegraphics[{width=.059\linewidth}]{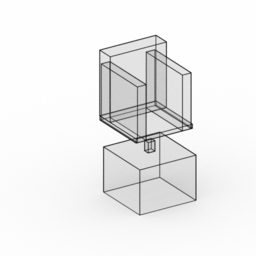} &
        \hspace{.75em}
        &
        \includegraphics[{width=.059\linewidth}]{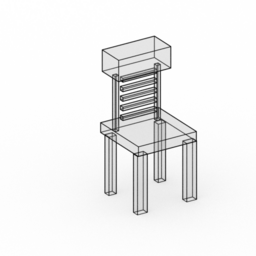} &
        \includegraphics[{width=.059\linewidth}]{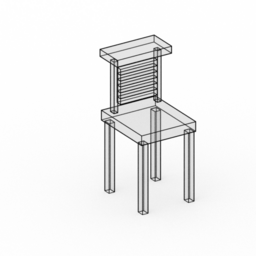} &
        \includegraphics[{width=.059\linewidth}]{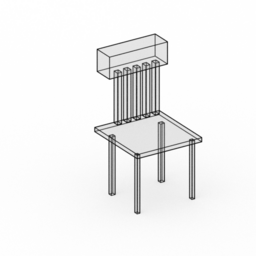} &
        \includegraphics[{width=.059\linewidth}]{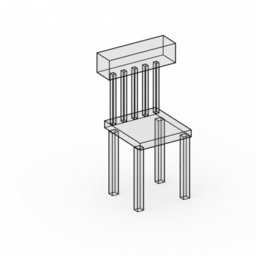} &
        \includegraphics[{width=.059\linewidth}]{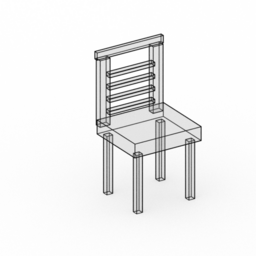} &
         \hspace{.75em}
        &
        \includegraphics[{width=.059\linewidth}]{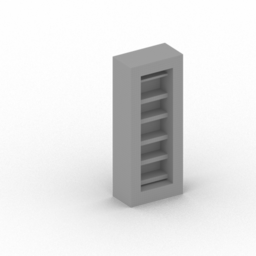} &
        \includegraphics[{width=.059\linewidth}]{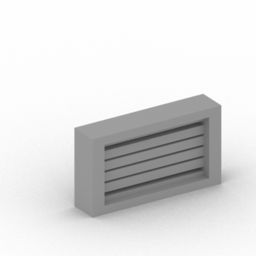} &
        \includegraphics[{width=.059\linewidth}]{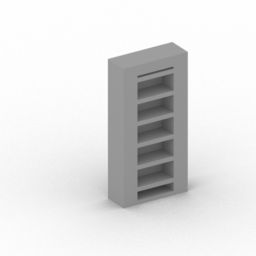} &
        \includegraphics[{width=.059\linewidth}]{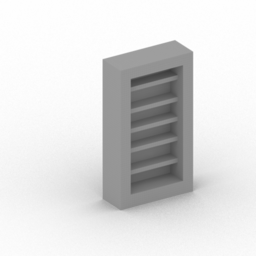} &
        \includegraphics[{width=.059\linewidth}]{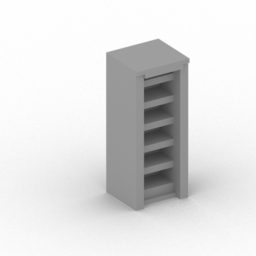} \\
        
    \end{tabular}
    \caption{We learn to infer Template Programs that capture input concepts (\textit{Inp}). Template Programs produce consistent concept parses (\textit{Seg}) and synthesize new generations (\textit{Gen}).
    Our framework flexibly extends across different visual domains and input representations.
    } 
    \label{fig:qual_main}
\end{figure*}

\textbf{Objective  }
Our inference procedure takes in a visual group~\VisGroup~and tries to find a Template Program~\TemplateGroup~whose instantiations~\ProgramGroup~best explain the group. 
We formalize this notion of \textit{best} with an objective composed of two terms (i) reconstruction error (under a domain-specific metric~\Metric) and  (ii) the description length difference between each~\program~and the~\Template~it originated from. Specifically, we try to minimize:
\begin{equation}
~\label{eq:obj}
\Objective = \lambda_1 * \sum_{\mathclap{(x,z) \in (X^G, Z^G)}}\Metric(x, E(z)) + \lambda_2 * {\sum\limits_{\program \in Z^G}}  |\program| - |TP^G| 
\end{equation}
In short, we search for Template Programs that encode as much commonality as possible while still producing instantiations that capture the visual input.

\section{Results}
\label{sec:results}

We validate the benefits of our method through comparisons with alternative approaches across three visual domains.
We describe the domains in Section~\ref{sec:res_domains} and our experimental design in Section~\ref{sec:res_exp_design}.
Next, we evaluate performance on downstream tasks: few-shot generative modeling (Section~\ref{sec:res_fsg}, Figure~\ref{fig:qual_main} \textit{Gen} rows) and parsing-based cosegmentation (Section~\ref{sec:res_coseg}, Figure~\ref{fig:qual_main} \textit{Seg} rows).
Finally, we discuss out-of-distribution generalization, method ablations, and additional capabilities of Template Programs in Section~\ref{sec:res_discussion}.

\subsection{Visual Domains}
\label{sec:res_domains}

We experiment over three visual domains that differ in input modality and concept groupings. 
We provide an overview of each domain here, and further information in Appendix~\ref{sec:app_domains}. 

\paragraph{2D Primitive Layouts}
We design a procedurally generated domain where concepts are represented with a layout of simple 2D colored primitives.
In addition to functions that move, scale, and color primitives, our DSL also contains simple symmetry functions (e.g. \texttt{REFLECT}, Fig.~\ref{fig:method}).
We hand-design 20 high-level meta-procedures that correspond with manufactured or organic concepts (e.g. cats or clocks). 
Each meta-procedure creates a distribution of concepts by expressing different combinations of four attributes, allowing us to produce 384 distinct concepts.
We divide these into 216 training-validation concepts and 168 testing concepts, where this split is designed to investigate out-of-distribution generalization performance (Section~\ref{sec:res_discussion}).

\textbf{Omniglot Characters  }
~\citet{bpl}~introduced the Omniglot dataset which contains handwritten characters from 50 languages.
These characters are split between a background set (964 characters) and a generalization set (659 characters), where each concept comes with 20 examples.
We use the background characters for training and validation, and test on the generalization characters.
Our DSL for drawing characters produces strokes by moving a virtual pen.
The pen moves at an angle, for varying distances, optionally bowing inwards or outwards. 
It can be lifted up or put down and has the option to back-track to previous positions.
As we are more interested in modeling stroke structure than physical handwriting dynamics, we adopt a simplified ink model compared with previous work: any pixel the pen passes through is filled completely.
 
\textbf{3D Shape Structures  }
Beyond 2D domains, we also run experiments on a dataset of 3D shapes.
Following past work, we use a structured part-based representation, where 3D shapes are modeled as a combination of primitives (i.e. cuboids)~\cite{EG2020STAR,PlankAssembly}. 
For our DSL, we use the ShapeAssembly modeling language~\cite{jones2020shapeAssembly}, which creates complex 3D shapes by instantiating cuboids and assembling them together through attachment and symmetry operators. 
We source 10,000 3D shape structures from the chair, table, and storage categories of PartNet~\cite{PartNet}, holding out 1000 of these for our test set.
We use the associated structural annotations in PartNet to identify groupings of these shapes that correspond to concepts that are more fine-grained than object category.
While we use annotations to partition the dataset into groups, 
our networks receive only a visual representation of each shape during training: either an unordered collection of primitives or a 3D voxel grid.

\begin{table*}[t!]
    \centering    
    \setlength{\tabcolsep}{1.5pt}
    \small
    \caption{Across multiple visual domains we quantitatively evaluate few-shot generation and co-segmentation performance. 
    Our method outperforms domain-general but task-specific alternatives, and is competitive against approaches that specialize for Omniglot.}
    \vspace{.5em}
    \begin{tabular}{@{}llccccccccccccccccccc@{}}

        \textbf{Domain} & &  \multicolumn{6}{c}{\rule[1.5pt]{5em}{0.5pt} \textbf{Omniglot} \rule[1.5pt]{5em}{0.5pt} } 
        & & \multicolumn{6}{c}{\rule[1.5pt]{5em}{0.5pt} \textbf{2D Layouts} \rule[1.5pt]{5em}{0.5pt}} 
        & &  \multicolumn{5}{c}{\rule[1.5pt]{3em}{0.5pt} \textbf{3D Shapes} \rule[1.5pt]{3em}{0.5pt}}\\

       \textbf{Task}&
       & \multicolumn{4}{c}{\textit{\rule[1.5pt]{2.5em}{0.5pt} Few-shot gen \rule[1.5pt]{2.5em}{0.5pt} }} 
       & & \multicolumn{1}{c}{\textit{Co-seg}} &
        & \multicolumn{4}{c}{\textit{\rule[1.5pt]{2.5em}{0.5pt}  Few-shot gen \rule[1.5pt]{2.5em}{0.5pt} }} 
        & & \multicolumn{1}{c}{\textit{Co-seg}} &
        & \multicolumn{3}{c}{\textit{\rule[1.5pt]{.5em}{0.5pt} Few-shot gen \rule[1.5pt]{.5em}{0.5pt} }} 
        & & \multicolumn{1}{c}{\textit{Co-seg}} 
        \\    
	&	\textbf{Method} \hspace{.2em} &  \textbf{FD}$\Downarrow$ & \textbf{Conf}$\Uparrow$  & \textbf{MMD}$\Downarrow$ & \textbf{Cov}$\Uparrow$  &  & \textbf{mIoU}$\Uparrow$  & & \textbf{FD}$\Downarrow$ & \textbf{Conf}$\Uparrow$  & \textbf{MMD}$\Downarrow$ & \textbf{Cov}$\Uparrow$  & & \textbf{mIoU}$\Uparrow$ & & \textbf{FD}$\Downarrow$ & \textbf{MMD}$\Downarrow$ & \textbf{Cov}$\Uparrow$  &  & \textbf{mIoU}$\Uparrow$  \\
		\midrule
		  \textit{Domain} \hspace{.2em} & BPL & 130 & 57.9 & 9.58 & \textbf{61.1}   &  & \textbf{79.9} & \hspace{.75em} & - & - & - & - && - & \hspace{.75em} & - & - & - && - \\
		\textit{Specific}& GNS & 123 & 55.0 & 9.47 & 58.1   & & 73.8 & & - & - & - & - && - & & - & - & - && - \\
        \midrule
		& FSDM  & 196 & 5.17 & 12.6 & 48.6  &  & - & & - & - & - & - && - & & - & - & - && - \\
		\textit{Task}& VHE & 139 & 2.46 & 10.4 & 52.0  & & - & & 81.9 & 59.0 & 8.06 & 22.4 && - & & - & - & - && - \\
		\textit{Specific}& arVHE & 137 & 12.3 & 10.2 & 55.8  & & - & & 45.3 & 77.0 & 6.34 & 45.1 && - & & 128 & 8.57 & 53.6  && -\\
		& BAE & - & - & - & - & & 34.3 & & - & - & - & - && 34.5 & & - & - & - && 53.2 \\
        \midrule
		& Ours & \textbf{115 }& \textbf{59.9} & \textbf{9.40} & 50.7  & & 78.7 & & \textbf{30.7} & \textbf{90.9} & \textbf{5.49} & \textbf{50.6} && \textbf{82.5 }& & \textbf{84.5} & \textbf{6.49} & \textbf{53.9} && \textbf{68.6} \\
        \bottomrule
    \end{tabular}
    \label{tab:main_table}
\end{table*}

\subsection{Experimental Design}
\label{sec:res_exp_design}

\paragraph{Networks}
We implement each autoregressive component of~\infnets~with Transformer decoder models that have 8 layers, 16 heads, and a hidden dimension of 256. 
We use causal attention masks with a prefix that contains conditioning information (see Section~\ref{sec:met_inf}, Appendix~\ref{sec:app_net_details}).
For the 2D layout and Omniglot domains we model our visual encoders with 2D CNNs that respectively take in RGB images of size 64x64 and binary images of size 28x28.
We train two different versions of~\infnets~for 3D shapes. 
When shapes are represented as an unordered collection of primitives (\textit{primitive soup}), we use a Transformer encoder with order-invariant positional encodings (Fig.~\ref{fig:qual_main}, left \& middle). 
We additionally explore using a 3D CNN that takes in a $64^3$ occupancy grid of voxels (Fig.~\ref{fig:qual_main}, right).
For each domain, we train~\infnets~with the procedure described in Sec.~\ref{sec:met_learn} until we reach convergence on the validation set (additional training details in Appendix~\ref{sec:app_train_details}).

\paragraph{Inference logic}
We infer Template Programs and their instantiations with a beam search. 
This algorithm has two parameters: $BM_{TP}$ controls the size of the beam used to find Template Programs under~\TemplateNet, while $BM_{z}$ controls the size of the beam used to find instantiated programs under~\ExpansionNet and~\ParamNet. 
This search concludes by evaluating each candidate under~\Objective, which requires a domain-specific reconstruction metric.
We use a color-based IoU for 2D layouts, an edge-based Chamfer distance for Omniglot, and either a primitive-matching score or IoU for 3D shapes depending on the input format (details in Appendix~\ref{sec:app_domains}). 
During fine-tuning, we set $BM_{TP}$ and $BM_{z}$ to 5 ($\sim$1 second for inference per input group). 
For evaluation tasks, we set $BM_{TP}$ to 40 and $BM_{z}$ to 10 ($\sim$20 seconds for inference per input group).

\paragraph{Comparison Conditions}
We compare how our method performs on concept-related tasks against alternative approaches.
For the Omniglot domain, we compare against the task-general but domain-specific \textit{BPL}~\cite{bpl} and \textit{GNS}~\cite{gns} methods.
Though they are designed to operate under one-shot paradigms, we adapt them for our task settings.
We also compare against alternatives that are domain-general but task-specific.
For few-shot generation, we compare against \textit{FSDM}~\cite{giannone2022fewshot} and \textit{VHE}~\cite{hewitt2018variational}.
These approaches both train deep generative networks that condition on a group of input images but use different generative models: VHE uses a VAE~\cite{kingma2014auto}, while FSDM uses diffusion~\cite{ho2020denoising}).
During our experiments, we found VAE training to be highly unstable, so we also introduced an autoregressive VHE variant: \textit{arVHE}.
Our arVHE model first tokenizes visual data (e.g. through vector-quantization~\cite{vqvae}) then learns an autoregressive model over this tokenization that is conditioned on groups of visual inputs.
For co-segmentation tasks, we compare against \textit{BAE-NET}~\cite{chen2019bae_net}.
BAE-NET forms consistent parses by training a parameter-constrained implicit network to solve an occupancy reconstruction task. 
Though this method is designed primarily for 2D and 3D shapes, we adapt it to create segmentations across all of our domains.
We provide additional details for all of our comparisons conditions in Appendix~\ref{sec:app_baseline}.

\subsection{Concept Few-shot generation}
\label{sec:res_fsg}

For few-shot generation, a method is given a set of examples from a concept as input and is tasked with producing new instances that demonstrate variety while maintaining concept membership.
Our method accomplishes this with a two step process: first we infer a Template Program that explains the input group, then we sample new instantiations from the Template Program.
To sample these instantiations, we use variants of our~\ExpansionNet~and~\ParamNet~that condition on a mean-pooled visual encoding of the input group (Appendix~\ref{sec:app_gcond_net}).
Across our three domains, we show examples of our method's few-shot generative capabilities in Figure~\ref{fig:qual_main}, \textit{Gen} rows.
Our method is able to capture input concepts and synthesize new outputs that demonstrate interesting variations while preserving concept identity.    

We present quantitative few-shot generation results in Table~\ref{tab:main_table} (details in App.~\ref{sec:app_exp_details}).
For each domain and test-set concept, we provide every method with a group of 5 visual inputs and ask it to synthesize 5 generations.
Comparing these generations to a reference set of held-out examples from the same concept, we compute the following metrics using the latent space of a domain-specific auto-encoder: 
Frechet Distance (\textit{FD}), 
Minimum Matching Distance (\textit{MMD}), and
Coverage (\textit{Cov}).
For the Omniglot and 2D layout domains, we also report class confidence~(\textit{Conf}), the average predicted probability of each generation being a member of the target class under a classifier trained on all domain concepts.

As demonstrated, our method vastly outperforms task-specific alternatives~(FSDM, VHE, arVHE) for few-shot generation. 
Over all domains, we find that our method scores much better along metrics that measure output concept consistency (\textit{Conf}) and fidelity to the reference set (\textit{FD}, \textit{MMD}), while maintaining reasonable output variability (\textit{Cov}). 
Moreover, our domain-general method is able to largely match, and even somewhat outperform, domain-specialized approaches (BPL, GNS) along measurements of concept consistency and fidelity to the reference set.

\begin{figure}[t]
    \centering
    \tiny
    \setlength{\tabcolsep}{.5pt}
    \begin{tabular}{cccccccccccc}
        {\raisebox{.3em}{\rotatebox{90}{Input}}} \hspace{.1em} &
        \includegraphics[{width=.085\linewidth}]{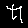} &
        \includegraphics[{width=.085\linewidth}]{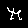} &
        \includegraphics[{width=.085\linewidth}]{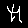} &
        \includegraphics[{width=.085\linewidth}]{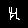} &
        \includegraphics[{width=.085\linewidth}]{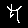} &
        \hspace{.3em} &
        \includegraphics[{width=.085\linewidth}]{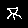} &
        \includegraphics[{width=.085\linewidth}]{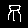} &
        \includegraphics[{width=.085\linewidth}]{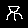} &
        \includegraphics[{width=.085\linewidth}]{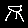} &
        \includegraphics[{width=.085\linewidth}]{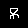} \\
                        {\raisebox{.35em}{\rotatebox{90}{Ours}}} \hspace{.1em} &
        \includegraphics[{width=.085\linewidth}]{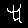} &
        \includegraphics[{width=.085\linewidth}]{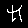} &
        \includegraphics[{width=.085\linewidth}]{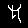} &
        \includegraphics[{width=.085\linewidth}]{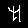} &
        \includegraphics[{width=.085\linewidth}]{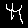} &
        \hspace{.3em} &
        \includegraphics[{width=.085\linewidth}]{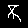} &
        \includegraphics[{width=.085\linewidth}]{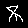} &
        \includegraphics[{width=.085\linewidth}]{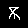} &
        \includegraphics[{width=.085\linewidth}]{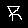} &
        \includegraphics[{width=.085\linewidth}]{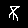} \\
        {\raisebox{.5em}{\rotatebox{90}{BPL}}} \hspace{.1em} &
        \includegraphics[{width=.085\linewidth}]{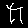} &
        \includegraphics[{width=.085\linewidth}]{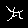} &
        \includegraphics[{width=.085\linewidth}]{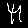} &
        \includegraphics[{width=.085\linewidth}]{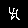} &
        \includegraphics[{width=.085\linewidth}]{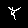} &
        \hspace{.3em} &
        \includegraphics[{width=.085\linewidth}]{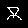} &
        \includegraphics[{width=.085\linewidth}]{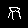} &
        \includegraphics[{width=.085\linewidth}]{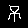} &
        \includegraphics[{width=.085\linewidth}]{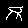} &
        \includegraphics[{width=.085\linewidth}]{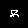} \\
            {\raisebox{.5em}{\rotatebox{90}{GNS}}} \hspace{.1em} &
        \includegraphics[{width=.085\linewidth}]{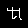} &
        \includegraphics[{width=.085\linewidth}]{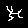} &
        \includegraphics[{width=.085\linewidth}]{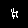} &
        \includegraphics[{width=.085\linewidth}]{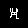} &
        \includegraphics[{width=.085\linewidth}]{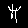} &
        \hspace{.3em} &
        \includegraphics[{width=.085\linewidth}]{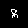} &
        \includegraphics[{width=.085\linewidth}]{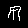} &
        \includegraphics[{width=.085\linewidth}]{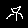} &
        \includegraphics[{width=.085\linewidth}]{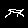} &
        \includegraphics[{width=.085\linewidth}]{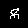} \\
        {\raisebox{.25em}{\rotatebox{90}{FSDM}}} \hspace{.1em} &
        \includegraphics[{width=.085\linewidth}]{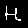} &
        \includegraphics[{width=.085\linewidth}]{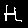} &
        \includegraphics[{width=.085\linewidth}]{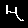} &
        \includegraphics[{width=.085\linewidth}]{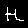} &
        \includegraphics[{width=.085\linewidth}]{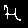} &
        \hspace{1em} &
        \includegraphics[{width=.085\linewidth}]{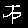} &
        \includegraphics[{width=.085\linewidth}]{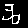} &
        \includegraphics[{width=.085\linewidth}]{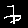} &
        \includegraphics[{width=.085\linewidth}]{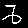} &
        \includegraphics[{width=.085\linewidth}]{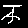} \\
        {\raisebox{.25em}{\rotatebox{90}{arVHE}}} \hspace{.1em} &
        \includegraphics[{width=.085\linewidth}]{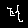} &
        \includegraphics[{width=.085\linewidth}]{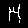} &
        \includegraphics[{width=.085\linewidth}]{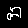} &
        \includegraphics[{width=.085\linewidth}]{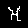} &
        \includegraphics[{width=.085\linewidth}]{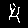} &
        \hspace{.3em} &
        \includegraphics[{width=.085\linewidth}]{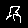} &
        \includegraphics[{width=.085\linewidth}]{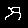} &
        \includegraphics[{width=.085\linewidth}]{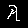} &
        \includegraphics[{width=.085\linewidth}]{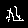} &
        \includegraphics[{width=.085\linewidth}]{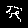} \\
    \end{tabular}
    \caption{\vspace{-2em} Comparing few-shot generations of Omniglot characters. \vspace{-1em}} 
    \label{fig:qual_omni}
\end{figure}

We visualize few-shot generation results for Omniglot characters in Figure~\ref{fig:qual_omni}.
While we again offer much improved performance over the task-specific alternatives, we note that the methods that specialize for Omniglot typically demonstrate a wider range of output variability, which confirms the trend we observe with the \textit{Cov} metric.
We hypothesize this difference is due to BPL and GNS learning priors over human stroke patterns (learning how people typically produce characters). In contrast, our method finds a Template Program attuned to the visual data present in the input group without regard for structured priors beyond the input DSL.

\paragraph{Perceptual Study}
To further investigate few-shot generative performance, we designed a two-alternative forced-choice perceptual study (Appendix \ref{sec:app_exp_percep_details}).
We recruited 20 participants, and presented a series of questions that compared generations from competing methods to the input group.
We report results for this study in Table~\ref{tab:percep}. 
For the Omniglot domain, we compared our method against our best performing task-general method (arVHE) and the domain-specific GNS method. 
We additionally compared our method against arVHE for the shape domain.
We observed that there was an overwhelming preference for our method compared with task-specific alternatives (our generations were preferred at rates of 94\% and 84\% against those produced by arVHE).
Even when our method was compared with GNS, we found participants had a slight preference for the few-shot generations our system produced, with 64\% preference rate.  
We point to this result as another strong indication of the impressive performance that our domain-general method is capable of achieving.

\subsection{Concept Co-segmentation }
\label{sec:res_coseg}

\begin{figure}[t]
    \centering
    \tiny
    \setlength{\tabcolsep}{.5pt}
    \begin{tabular}{cccccc}

        \raisebox{3.0em}{Input} \hspace{.25em} &
        \includegraphics[{width=.16\linewidth}]{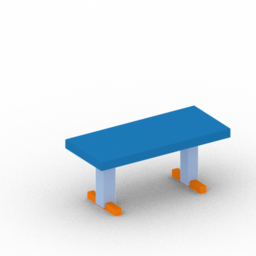} &
        \includegraphics[{width=.16\linewidth}]{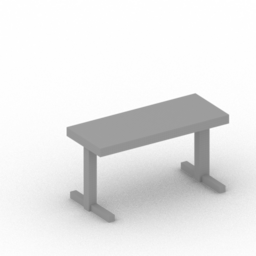} &
        \includegraphics[{width=.16\linewidth}]{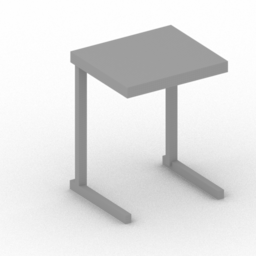} &
        \includegraphics[{width=.16\linewidth}]{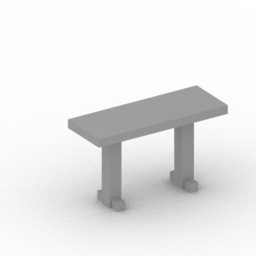} &
        \includegraphics[{width=.16\linewidth}]{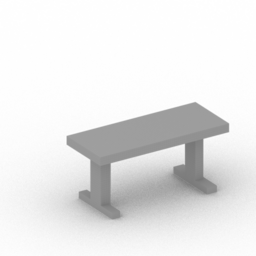} \\
        \raisebox{3.0em}{BAE} \hspace{.25em} &
         &
        \includegraphics[{width=.16\linewidth}]{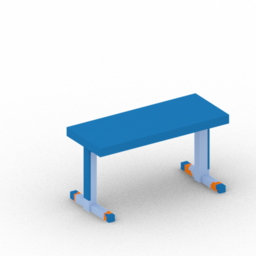} &
        \includegraphics[{width=.16\linewidth}]{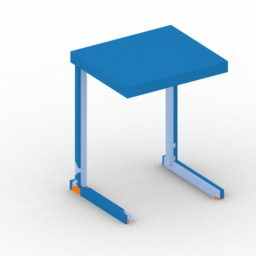} &
        \includegraphics[{width=.16\linewidth}]{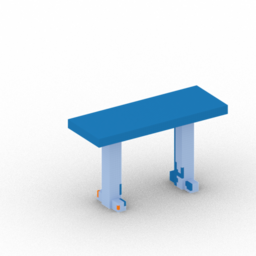} &
        \includegraphics[{width=.16\linewidth}]{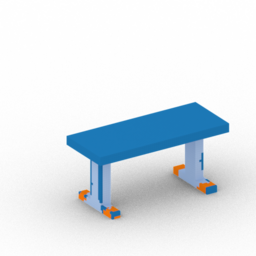}  \\
        \raisebox{3.0em}{Ours} \hspace{.25em} &
         &
        \includegraphics[{width=.16\linewidth}]{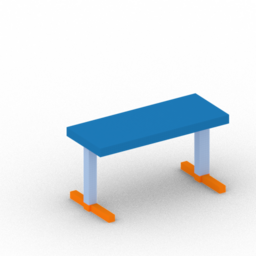} &
        \includegraphics[{width=.16\linewidth}]{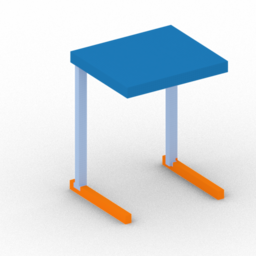} &
        \includegraphics[{width=.16\linewidth}]{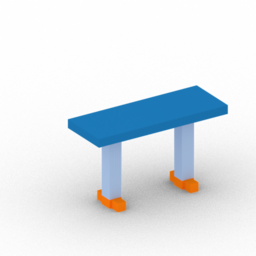} &
        \includegraphics[{width=.16\linewidth}]{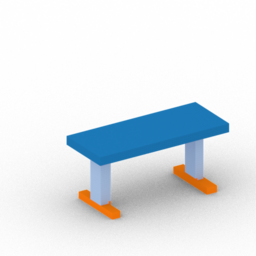} \\
         \raisebox{3.0em}{GT} \hspace{.25em} &
         &
        \includegraphics[{width=.16\linewidth}]{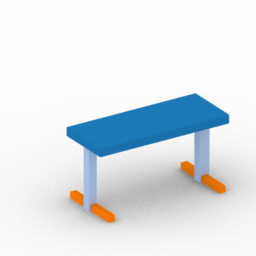} &
        \includegraphics[{width=.16\linewidth}]{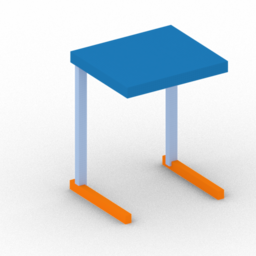} &
        \includegraphics[{width=.16\linewidth}]{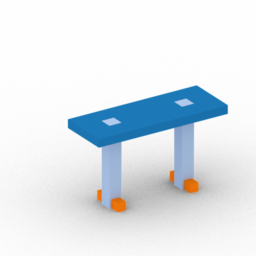} &
        \includegraphics[{width=.16\linewidth}]{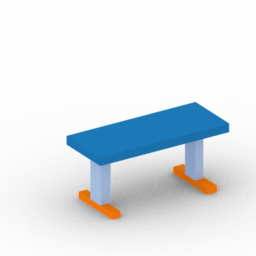} \\
    \end{tabular}
    \caption{We compare co-segmentations produced from voxelized shapes (\textit{Input}) to ground-truth annotations (\textit{GT}) \vspace{-1em}} 
    \label{fig:qual_shape}
\end{figure}

Our method also natively supports co-analysis tasks. 
When we infer a Template Program and instantiations that explain an input visual group, we can use the shared structure of the Template Program to parse the group members in a consistent fashion.
We visualize this capability in the \textit{Seg} rows of~Figure~\ref{fig:qual_main}.
This consistent parsing allows us to perform a co-segmentation task:
given an input visual group, where exactly one member of the group has a labeled segmentation, our goal is to propagate this labeling to the other group members.
We provide further details in Appendix~\ref{sec:app_exp_coseg_details}.

We compare how our method does on co-segmentation tasks across domains.
Our main comparison is against BAE-NET~\cite{chen2019bae_net}, which is designed specifically for this task.
For Omniglot, BPL and GNS can also perform this task by parsing visual inputs to ordered strokes. 
We report results of our experiments in Table~\ref{tab:main_table}. 
We evaluate performance with a mean intersection over union metric (\textit{mIoU}) that measures how closely the output segmentation predictions match the ground-truth labelings.
Despite the fact that our method never trains on human stroke data,
we achieve a better mIoU on this co-segmentation task compared with GNS, and nearly match the metric value achieved by BPL. 
Though our output co-segmentations are less structured compared with the ordered stroke parses BPL and GNS can produce, we are encouraged by our method's performance in this task.
For our 3D shapes domain, as BAE-NET was originally designed to operate over voxels, our comparisons against it use a variant of our method that also takes in voxel inputs.
We visualize an example co-segmentation of each method in Figure~\ref{fig:qual_shape}.
Across domains and input modalities, we find that we outperform BAE-NET for this  task.

\subsection{Discussion}
\label{sec:res_discussion}

\paragraph{Out-of-distribution generalization}
Different domains require different levels of generalization.
For instance, in the Omniglot dataset there is no alphabet overlap between train and test characters, so strong generalization capabilities are required for each test concept.
As we procedurally generated the 2D layout domain, we are able to control and evaluate the level of out-of-distribution generalization required for each test-set concept. 
We consider three settings.
\textit{Easy} concepts have a new combination of attributes, but each attribute has been seen before (e.g. chair back, top-left of Fig.~\ref{fig:qual_main}).
\textit{Medium} concepts have a new attribute not seen during training (e.g. double-sided leaves, top-middle of Fig.~\ref{fig:qual_main}).
\textit{Hard} concepts are from a meta-procedure that was not used at all during training (e.g. turtles, top-right of Fig.~\ref{fig:qual_main}).
We find that while our method does become worse when evaluated on more difficult concepts, its performance remains more consistent compared with alternative approaches.
We explore this phenomenon further in Appendix~\ref{sec:app_layout_gen}.

\paragraph{Ablations}
We consider the effect of different design decisions on our method with an ablation study.
We provide the details of this study and quantitative results in Appendix~\ref{sec:app_ablation}.
We find that our bootstrapped finetuning process is critical to adapting networks pretrained on synthetic data towards a target dataset of interest.
We validate that our scheme of allowing the Template Program to capture parametric relationships improves performance on downstream tasks.
Finally we compare our three step inference approach (\Template~$\rightarrow$~\Expansion~$\rightarrow$~\program) against a two step alternative where each~\program~is predicted directly from the \Template. In this comparison, we find that our formulation, which allows the ParamNet to attend over the complete expression tree, outperforms this alternative formulation.

\paragraph{Unconditional Concept Generation}
Though we mainly evaluate our method on few-shot generation and co-segmentation, these are not the only concept-related tasks our framework can support.
For Omniglot, we explore how our approach can be used for unconditional concept generation.
In fact, this is a task we naturally solve as part of our fine-tuning procedure: the wake-sleep component of each training loop uses an unconditional generative model to sample \textit{Template Programs} that represent new concepts.
We visualize some of these generations in Figure~\ref{fig:uncond_con_gen}.

\begin{table}[t]
    \centering
    \caption{Perceptual study results evaluating few-shot generation performance. Our method is greatly preferred over task-specific alternatives and slightly preferred over domain-specific alternatives.}
    \vspace{1em}
    \begin{tabular}{lcccc} 
        \textbf{Domain} & \multicolumn{2}{c}{\rule[1.5pt]{2em}{0.5pt}\textit{Omniglot}\rule[1.5pt]{2em}{0.5pt}} 
        & & \textit{3D Shapes} \\
         & \textbf{arVHE} & \textbf{GNS} & & \textbf{arVHE}  \\
        \midrule
        \textbf{Ours vs.} & 94\% & 64\% &&  84\% \\
         
    \end{tabular}
    \label{tab:percep}
\end{table}

\section{Conclusion}
\label{sec:conclusion}

We presented the \textit{Template Programs} framework: a neurosymbolic method that learns to capture visual concepts with structured symbolic objects.
We demonstrated that our method flexibly learns to infer Template Programs across multiple visual domains: 2D primitive layouts, Omniglot characters, and 3D shape structures.
Our approach supports multiple downstream tasks of interest, such as few-shot generation and co-segmentation.
On these tasks, we achieve superior performance over other domain-general,  \textit{task-specific} alternatives, and find that we match, and in some cases slightly outperform, \textit{domain-specific}, task-general alternatives for the limited areas where they exist.

There are a number of directions we would like to investigate in future work. 
So far, we have only considered visual input groups of size 5, but this constraint can be relaxed by changing the conditioning information presented to the TemplateNet. 
More generally, one could train this network over input groups of varying sizes (i.e through random masking of visual encodings), to achieve the most inference time flexibility.
Further, while we allow our Template Programs to capture parametric relations, the kinds of relations we have so far investigated are fairly simple: static variable assignment and variable re-use. 
For continuous variables especially, it would be interesting to consider learning more complex relations.
These could be declarative (e.g. closed-formula equations that operate over other parameters) or could even describe distributions (e.g. by parameterizing a Gaussian).
Our \textit{Template Programs} framework offers a promising step forward towards general concept learning, and looking ahead, we are excited to see how it can be adapted to an even broader array of tasks and domains.

\section*{Acknowledgments}

We would like to thank the anonymous reviewers for their helpful suggestions and all of our perceptual study particpants for their time.
Renderings of 3D shapes were produced using the Blender Cycles renderer. 
This work was funded in parts by NSF award \#1941808 and a Brown University Presidential Fellowship.
Daniel Ritchie is an advisor to Geopipe and owns equity in the company. 
Geopipe is a start-up that is developing 3D technology to build immersive virtual copies of the real world with applications in various fields, including games and architecture.
Part of this work was done while R. Kenny Jones was an intern at Adobe Research.

\section*{Impact Statement}

We propose a domain and task general system that learns to capture visual concepts programmatically. 
Our system can be used for concept related tasks, such as few-shot generative modeling and co-analysis.
While we don't envision direct negative impacts of our system, generative models can have both positive and negative effects dependent on use-case.
Our system learns generative models for specific domains, so data privacy and legality considerations are unlikely to affect our method.
While any progress on few-shot generative modeling could potentially lead to harmful actions through nefarious actors, we believe this outcome highly unlikely for our model as we are principally concerned with modeling structured visual data that can be represented programmatically. 
These domains have much less potential for abuse compared with generative models that learn over distributions of people, for example.

\bibliography{main}

\begin{thebibliography}{55}
\providecommand{\natexlab}[1]{#1}
\providecommand{\url}[1]{\texttt{#1}}
\expandafter\ifx\csname urlstyle\endcsname\relax
  \providecommand{\doi}[1]{doi: #1}\else
  \providecommand{\doi}{doi: \begingroup \urlstyle{rm}\Url}\fi

\bibitem[Achlioptas et~al.(2018)Achlioptas, Diamanti, Mitliagkas, and
  Guibas]{achlioptas2018learning}
Achlioptas, P., Diamanti, O., Mitliagkas, I., and Guibas, L.
\newblock Learning representations and generative models for 3d point clouds,
  2018.

\bibitem[Chaudhuri et~al.(2020)Chaudhuri, Ritchie, Wu, Xu, and
  Zhang]{EG2020STAR}
Chaudhuri, S., Ritchie, D., Wu, J., Xu, K., and Zhang, H.
\newblock {Learning Generative Models of 3D Structures}.
\newblock \emph{Computer Graphics Forum}, 2020.
\newblock ISSN 1467-8659.
\newblock \doi{10.1111/cgf.14020}.

\bibitem[Chen et~al.(2019)Chen, Yin, Fisher, Chaudhuri, and
  Zhang]{chen2019bae_net}
Chen, Z., Yin, K., Fisher, M., Chaudhuri, S., and Zhang, H.
\newblock Bae-net: Branched autoencoder for shape co-segmentation.
\newblock \emph{Proceedings of International Conference on Computer Vision
  (ICCV)}, 2019.

\bibitem[{Demir} et~al.(2016){Demir}, {Aliaga}, and
  {Benes}]{InverseProceduralArchitecture}
{Demir}, I., {Aliaga}, D.~G., and {Benes}, B.
\newblock Proceduralization for editing 3d architectural models.
\newblock In \emph{2016 Fourth International Conference on 3D Vision (3DV)},
  2016.

\bibitem[Du et~al.(2018)Du, Inala, Pu, Spielberg, Schulz, Rus, Solar-Lezama,
  and Matusik]{du2018inversecsg}
Du, T., Inala, J.~P., Pu, Y., Spielberg, A., Schulz, A., Rus, D., Solar-Lezama,
  A., and Matusik, W.
\newblock Inversecsg: automatic conversion of {3D} models to csg trees.
\newblock In \emph{Annual Conference on Computer Graphics and Interactive
  Techniques Asia (SIGGRAPH Asia)}. ACM, 2018.

\bibitem[Edwards \& Storkey(2017)Edwards and Storkey]{neuralstat}
Edwards, H. and Storkey, A.
\newblock Towards a neural statistician.
\newblock In \emph{5th International Conference on Learning Representations
  (ICLR 2017)}, pp.\  1--13, April 2017.
\newblock 5th International Conference on Learning Representations, ICLR 2017 ;
  Conference date: 24-04-2017 Through 26-04-2017.

\bibitem[Ellis et~al.(2018)Ellis, Ritchie, Solar-Lezama, and
  Tenenbaum]{handdrawn}
Ellis, K., Ritchie, D., Solar-Lezama, A., and Tenenbaum, J.~B.
\newblock Learning to infer graphics programs from hand-drawn images.
\newblock NIPS'18, pp.\  6062–6071, Red Hook, NY, USA, 2018. Curran
  Associates Inc.

\bibitem[Ellis et~al.(2019)Ellis, Nye, Pu, Sosa, Tenenbaum, and
  Solar-Lezama]{ellis2019write}
Ellis, K., Nye, M., Pu, Y., Sosa, F., Tenenbaum, J., and Solar-Lezama, A.
\newblock Write, execute, assess: Program synthesis with a repl.
\newblock In \emph{Advances in Neural Information Processing Systems
  (NeurIPS)}, 2019.

\bibitem[Ellis et~al.(2021)Ellis, Wong, Nye, Sabl{\'e}-Meyer, Morales, Hewitt,
  Cary, Solar-Lezama, and Tenenbaum]{DreamCoder}
Ellis, K., Wong, C., Nye, M., Sabl{\'e}-Meyer, M., Morales, L., Hewitt, L.,
  Cary, L., Solar-Lezama, A., and Tenenbaum, J.~B.
\newblock {DreamCoder}: Bootstrapping inductive program synthesis with
  wake-sleep library learning.
\newblock In \emph{ACM SIGPLAN International Symposium on New Ideas, New
  Paradigms, and Reflections on Programming and Software (SIGPLAN)}, pp.\
  835--850, 2021.

\bibitem[Ester et~al.(1996)Ester, Kriegel, Sander, and Xu]{dbscan}
Ester, M., Kriegel, H.-P., Sander, J., and Xu, X.
\newblock A density-based algorithm for discovering clusters in large spatial
  databases with noise.
\newblock In \emph{Proceedings of the Second International Conference on
  Knowledge Discovery and Data Mining}, KDD'96, pp.\  226–231. AAAI Press,
  1996.

\bibitem[Feinman \& Lake(2021)Feinman and Lake]{gns}
Feinman, R. and Lake, B.~M.
\newblock Learning task-general representations with generative neuro-symbolic
  modeling.
\newblock In \emph{International Conference on Learning Representations}, 2021.

\bibitem[Finn et~al.(2017)Finn, Abbeel, and Levine]{maml}
Finn, C., Abbeel, P., and Levine, S.
\newblock Model-agnostic meta-learning for fast adaptation of deep networks.
\newblock In Precup, D. and Teh, Y.~W. (eds.), \emph{Proceedings of the 34th
  International Conference on Machine Learning}, volume~70 of \emph{Proceedings
  of Machine Learning Research}, pp.\  1126--1135. PMLR, 06--11 Aug 2017.

\bibitem[Ganeshan et~al.(2023)Ganeshan, Jones, and Ritchie]{ganeshan2023coref}
Ganeshan, A., Jones, R.~K., and Ritchie, D.
\newblock Improving unsupervised visual program inference with code rewriting
  families.
\newblock In \emph{Proceedings of the International Conference on Computer
  Vision ({ICCV})}, 2023.

\bibitem[Ganin et~al.(2018)Ganin, Kulkarni, Babuschkin, Eslami, and
  Vinyals]{spiral}
Ganin, Y., Kulkarni, T., Babuschkin, I., Eslami, S. M.~A., and Vinyals, O.
\newblock Synthesizing programs for images using reinforced adversarial
  learning.
\newblock \emph{CoRR}, abs/1804.01118, 2018.

\bibitem[Giannone et~al.(2022)Giannone, Nielsen, and
  Winther]{giannone2022fewshot}
Giannone, G., Nielsen, D., and Winther, O.
\newblock Few-shot diffusion models, 2022.

\bibitem[Guo et~al.(2020)Guo, Jiang, Benes, Deussen, Zhang, Lischinski, and
  Huang]{guo2020inverse}
Guo, J., Jiang, H., Benes, B., Deussen, O., Zhang, X., Lischinski, D., and
  Huang, H.
\newblock Inverse procedural modeling of branching structures by inferring
  l-systems.
\newblock \emph{ACM Transactions on Graphics (TOG)}, 39\penalty0 (5):\penalty0
  1--13, 2020.

\bibitem[Heusel et~al.(2017)Heusel, Ramsauer, Unterthiner, Nessler, and
  Hochreiter]{FrechetInceptionDistance}
Heusel, M., Ramsauer, H., Unterthiner, T., Nessler, B., and Hochreiter, S.
\newblock Gans trained by a two time-scale update rule converge to a local nash
  equilibrium.
\newblock In \emph{Advances in Neural Information Processing Systems
  (NeurIPS)}, 2017.

\bibitem[Hewitt et~al.()Hewitt, Le, and Tenenbaum]{MemoizedWakeSleep}
Hewitt, L.~B., Le, T.~A., and Tenenbaum, J.~B.
\newblock Learning to learn generative programs with memoised wake-sleep.
\newblock In \emph{Uncertainty in Artificial Intelligence}.

\bibitem[Hewitt et~al.(2018)Hewitt, Nye, Gane, Jaakkola, and
  Tenenbaum]{hewitt2018variational}
Hewitt, L.~B., Nye, M.~I., Gane, A., Jaakkola, T., and Tenenbaum, J.~B.
\newblock The variational homoencoder: Learning to learn high capacity
  generative models from few examples, 2018.

\bibitem[Ho et~al.(2020)Ho, Jain, and Abbeel]{ho2020denoising}
Ho, J., Jain, A., and Abbeel, P.
\newblock Denoising diffusion probabilistic models.
\newblock \emph{arXiv preprint arxiv:2006.11239}, 2020.

\bibitem[Hu et~al.(2023)Hu, Zheng, Zhang, Yuan, Yin, and Zhou]{PlankAssembly}
Hu, W., Zheng, J., Zhang, Z., Yuan, X., Yin, J., and Zhou, Z.
\newblock Plankassembly: Robust 3d reconstruction from three orthographic views
  with learnt shape programs.
\newblock In \emph{ICCV}, 2023.

\bibitem[Hwang et~al.(2011)Hwang, Stuhlm\"{u}ller, and
  Goodman]{BayesianProgramMerging}
Hwang, I., Stuhlm\"{u}ller, A., and Goodman, N.~D.
\newblock {Inducing Probabilistic Programs by Bayesian Program Merging}.
\newblock \emph{CoRR}, arXiv:1110.5667, 2011.

\bibitem[Jones et~al.(2020)Jones, Barton, Xu, Wang, Jiang, Guerrero, Mitra, and
  Ritchie]{jones2020shapeAssembly}
Jones, R.~K., Barton, T., Xu, X., Wang, K., Jiang, E., Guerrero, P., Mitra,
  N.~J., and Ritchie, D.
\newblock Shapeassembly: Learning to generate programs for 3d shape structure
  synthesis.
\newblock \emph{ACM Transactions on Graphics (TOG)}, 39\penalty0 (6), 2020.

\bibitem[Jones et~al.(2022)Jones, Walke, and Ritchie]{jones2022PLAD}
Jones, R.~K., Walke, H., and Ritchie, D.
\newblock Plad: Learning to infer shape programs with pseudo-labels and
  approximate distributions.
\newblock \emph{The IEEE Conference on Computer Vision and Pattern Recognition
  (CVPR)}, 2022.

\bibitem[Kingma \& Ba(2015)Kingma and Ba]{Kingma2014AdamAM}
Kingma, D. and Ba, J.
\newblock Adam: A method for stochastic optimization.
\newblock In \emph{International Conference on Learning Representations
  (ICLR)}, 2015.

\bibitem[Kingma \& Welling(2014)Kingma and Welling]{kingma2014auto}
Kingma, D.~P. and Welling, M.
\newblock {Auto-{Encoding} Variational Bayes}.
\newblock In \emph{International Conference on Learning Representations
  (ICLR)}, 2014.

\bibitem[Kuhn(1955)]{kuhn1955hungarian}
Kuhn, H.~W.
\newblock The hungarian method for the assignment problem.
\newblock \emph{Naval research logistics quarterly}, 2\penalty0 (1-2):\penalty0
  83--97, 1955.

\bibitem[Lake et~al.(2015)Lake, Salakhutdinov, and Tenenbaum]{bpl}
Lake, B.~M., Salakhutdinov, R., and Tenenbaum, J.~B.
\newblock Human-level concept learning through probabilistic program induction.
\newblock \emph{Science}, 350\penalty0 (6266):\penalty0 1332--1338, 2015.
\newblock \doi{10.1126/science.aab3050}.

\bibitem[Lake et~al.(2019{\natexlab{a}})Lake, Salakhutdinov, and
  Tenenbaum]{omniglot}
Lake, B.~M., Salakhutdinov, R., and Tenenbaum, J.~B.
\newblock The omniglot challenge: a 3-year progress report.
\newblock \emph{Current Opinion in Behavioral Sciences}, 29:\penalty0 97--104,
  2019{\natexlab{a}}.
\newblock ISSN 2352-1546.
\newblock \doi{https://doi.org/10.1016/j.cobeha.2019.04.007}.
\newblock Artificial Intelligence.

\bibitem[Lake et~al.(2019{\natexlab{b}})Lake, Salakhutdinov, and
  Tenenbaum]{omniprog}
Lake, B.~M., Salakhutdinov, R., and Tenenbaum, J.~B.
\newblock The omniglot challenge: a 3-year progress report.
\newblock \emph{Current Opinion in Behavioral Sciences}, 29:\penalty0 97--104,
  2019{\natexlab{b}}.
\newblock ISSN 2352-1546.
\newblock \doi{https://doi.org/10.1016/j.cobeha.2019.04.007}.
\newblock Artificial Intelligence.

\bibitem[Liang et~al.(2022)Liang, Tenenbaum, Le, and N]{dood}
Liang, Y., Tenenbaum, J., Le, T.~A., and N, S.
\newblock Drawing out of distribution with neuro-symbolic generative models.
\newblock In Koyejo, S., Mohamed, S., Agarwal, A., Belgrave, D., Cho, K., and
  Oh, A. (eds.), \emph{Advances in Neural Information Processing Systems},
  volume~35, pp.\  15244--15254. Curran Associates, Inc., 2022.

\bibitem[Martinovic \& Van~Gool(2013)Martinovic and Van~Gool]{Martinovic_prog}
Martinovic, A. and Van~Gool, L.
\newblock Bayesian grammar learning for inverse procedural modeling.
\newblock In \emph{Proceedings of the 2013 IEEE Conference on Computer Vision
  and Pattern Recognition}, CVPR '13, pp.\  201–208, USA, 2013. IEEE Computer
  Society.
\newblock ISBN 9780769549897.
\newblock \doi{10.1109/CVPR.2013.33}.

\bibitem[Mo et~al.(2019)Mo, Zhu, Chang, Yi, Tripathi, Guibas, and Su]{PartNet}
Mo, K., Zhu, S., Chang, A.~X., Yi, L., Tripathi, S., Guibas, L.~J., and Su, H.
\newblock {PartNet}: A large-scale benchmark for fine-grained and hierarchical
  part-level {3D} object understanding.
\newblock In \emph{The IEEE Conference on Computer Vision and Pattern
  Recognition (CVPR)}, June 2019.

\bibitem[Murphy(2004)]{murphy2004big}
Murphy, G.
\newblock \emph{The big book of concepts}.
\newblock MIT press, 2004.

\bibitem[Nishida et~al.(2016)Nishida, Garcia-Dorado, Aliaga, Benes, and
  Bousseau]{nishida_proc}
Nishida, G., Garcia-Dorado, I., Aliaga, D.~G., Benes, B., and Bousseau, A.
\newblock Interactive sketching of urban procedural models.
\newblock \emph{ACM Trans. Graph.}, 35\penalty0 (4), jul 2016.
\newblock ISSN 0730-0301.
\newblock \doi{10.1145/2897824.2925951}.

\bibitem[Nishida et~al.(2018)Nishida, Bousseau, and G.~Aliaga]{NBG18}
Nishida, G., Bousseau, A., and G.~Aliaga, D.
\newblock Procedural modeling of a building from a single image.
\newblock \emph{Computer Graphics Forum (Proceedings of the Eurographics
  conference)}, 2018.

\bibitem[Paszke et~al.(2017)Paszke, Gross, Chintala, Chanan, Yang, DeVito, Lin,
  Desmaison, Antiga, and Lerer]{paszke2017automatic}
Paszke, A., Gross, S., Chintala, S., Chanan, G., Yang, E., DeVito, Z., Lin, Z.,
  Desmaison, A., Antiga, L., and Lerer, A.
\newblock Automatic differentiation in pytorch.
\newblock In \emph{Advances in Neural Information Processing Systems
  (NeurIPS)}, 2017.

\bibitem[Raffel et~al.(2019)Raffel, Shazeer, Roberts, Lee, Narang, Matena,
  Zhou, Li, and Liu]{t5paper}
Raffel, C., Shazeer, N., Roberts, A., Lee, K., Narang, S., Matena, M., Zhou,
  Y., Li, W., and Liu, P.~J.
\newblock Exploring the limits of transfer learning with a unified text-to-text
  transformer.
\newblock \emph{CoRR}, abs/1910.10683, 2019.

\bibitem[Rezende et~al.(2016)Rezende, Mohamed, Danihelka, Gregor, and
  Wierstra]{rezende2016oneshot}
Rezende, D.~J., Mohamed, S., Danihelka, I., Gregor, K., and Wierstra, D.
\newblock One-shot generalization in deep generative models, 2016.

\bibitem[Ritchie et~al.(2018)Ritchie, Jobalia, and Thomas]{ProcmodLearn}
Ritchie, D., Jobalia, S., and Thomas, A.
\newblock Example-based authoring of procedural modeling programs with
  structural and continuous variability.
\newblock In \emph{EUROGRAPHICS}, 2018.

\bibitem[Ritchie et~al.(2023)Ritchie, Guerrero, Jones, Mitra, Schulz, Willis,
  and Wu]{NcsgSTAR}
Ritchie, D., Guerrero, P., Jones, R.~K., Mitra, N.~J., Schulz, A., Willis, K.
  l. D.~D., and Wu, J.
\newblock {Neurosymbolic Models for Computer Graphics}.
\newblock \emph{Computer Graphics Forum}, 2023.

\bibitem[Ruiz et~al.(2022)Ruiz, Li, Jampani, Pritch, Rubinstein, and
  Aberman]{ruiz2022dreambooth}
Ruiz, N., Li, Y., Jampani, V., Pritch, Y., Rubinstein, M., and Aberman, K.
\newblock Dreambooth: Fine tuning text-to-image diffusion models for
  subject-driven generation.
\newblock 2022.

\bibitem[Rybkin et~al.(2020)Rybkin, Daniilidis, and Levine]{sigmavae}
Rybkin, O., Daniilidis, K., and Levine, S.
\newblock Simple and effective vae training with calibrated decoders, 2020.

\bibitem[Sharma et~al.(2018)Sharma, Goyal, Liu, Kalogerakis, and
  Maji]{sharma2018csgnet}
Sharma, G., Goyal, R., Liu, D., Kalogerakis, E., and Maji, S.
\newblock {{CSGNet}: Neural Shape Parser for Constructive Solid Geometry}.
\newblock In \emph{IEEE Conference on Computer Vision and Pattern Recognition
  (CVPR)}, 2018.

\bibitem[Snell et~al.(2017)Snell, Swersky, and Zemel]{protonets}
Snell, J., Swersky, K., and Zemel, R.
\newblock Prototypical networks for few-shot learning.
\newblock In Guyon, I., Luxburg, U.~V., Bengio, S., Wallach, H., Fergus, R.,
  Vishwanathan, S., and Garnett, R. (eds.), \emph{Advances in Neural
  Information Processing Systems}, volume~30. Curran Associates, Inc., 2017.

\bibitem[Solar-Lezama(2008)]{solar2008program}
Solar-Lezama, A.
\newblock \emph{Program Synthesis by Sketching}.
\newblock PhD thesis, UNIVERSITY OF CALIFORNIA, BERKELEY, 2008.

\bibitem[Stava et~al.(2010)Stava, Benes, Mech, Aliaga, and
  Kristof]{InverseLSystems}
Stava, O., Benes, B., Mech, R., Aliaga, D.~G., and Kristof, P.
\newblock Inverse procedural modeling by automatic generation of l-systems.
\newblock \emph{Computer Graphics Forum (CGF)}, 29:\penalty0 665--674, 2010.

\bibitem[Stuhlmuller et~al.(2010)Stuhlmuller, Tenenbaum, and
  Goodman]{stuhlmuller2010learning}
Stuhlmuller, A., Tenenbaum, J.~B., and Goodman, N.~D.
\newblock Learning structured generative concepts.
\newblock Cognitive Science Society, 2010.

\bibitem[Tenenbaum(2018)]{likepeople}
Tenenbaum, J.
\newblock Building machines that learn and think like people.
\newblock In \emph{Proceedings of the 17th International Conference on
  Autonomous Agents and MultiAgent Systems}, AAMAS '18, pp.\ ~5, Richland, SC,
  2018. International Foundation for Autonomous Agents and Multiagent Systems.

\bibitem[Tian et~al.(2019)Tian, Luo, Sun, Ellis, Freeman, Tenenbaum, and
  Wu]{tian2019learning}
Tian, Y., Luo, A., Sun, X., Ellis, K., Freeman, W.~T., Tenenbaum, J.~B., and
  Wu, J.
\newblock {Learning to Infer and Execute {3D} Shape Programs}.
\newblock In \emph{International Conference on Learning Representations
  (ICLR)}, 2019.

\bibitem[van~den Oord et~al.(2017)van~den Oord, Vinyals, and
  kavukcuoglu]{vqvae}
van~den Oord, A., Vinyals, O., and kavukcuoglu, k.
\newblock Neural discrete representation learning.
\newblock In Guyon, I., Luxburg, U.~V., Bengio, S., Wallach, H., Fergus, R.,
  Vishwanathan, S., and Garnett, R. (eds.), \emph{Advances in Neural
  Information Processing Systems}, volume~30. Curran Associates, Inc., 2017.

\bibitem[Vaswani et~al.(2017)Vaswani, Shazeer, Parmar, Uszkoreit, Jones, Gomez,
  Kaiser, and Polosukhin]{att_is_all}
Vaswani, A., Shazeer, N., Parmar, N., Uszkoreit, J., Jones, L., Gomez, A.~N.,
  Kaiser, L.~u., and Polosukhin, I.
\newblock Attention is all you need.
\newblock In Guyon, I., Luxburg, U.~V., Bengio, S., Wallach, H., Fergus, R.,
  Vishwanathan, S., and Garnett, R. (eds.), \emph{Advances in Neural
  Information Processing Systems}, volume~30. Curran Associates, Inc., 2017.

\bibitem[Vinyals et~al.(2016)Vinyals, Blundell, Lillicrap, kavukcuoglu, and
  Wierstra]{matchnets}
Vinyals, O., Blundell, C., Lillicrap, T., kavukcuoglu, k., and Wierstra, D.
\newblock Matching networks for one shot learning.
\newblock In Lee, D., Sugiyama, M., Luxburg, U., Guyon, I., and Garnett, R.
  (eds.), \emph{Advances in Neural Information Processing Systems}, volume~29.
  Curran Associates, Inc., 2016.

\bibitem[Xu et~al.(2021)Xu, Peng, Cheng, Willis, and Ritchie]{zoneGraphs}
Xu, X., Peng, W., Cheng, C.-Y., Willis, K. D.~D., and Ritchie, D.
\newblock Inferring {CAD} modeling sequences using zone graphs.
\newblock In \emph{IEEE Conference on Computer Vision and Pattern Recognition
  (CVPR)}, 2021.

\bibitem[Xu et~al.(2022)Xu, Willis, Lambourne, Cheng, Jayaraman, and
  Furukawa]{xu2022skexgen}
Xu, X., Willis, K.~D., Lambourne, J.~G., Cheng, C.-Y., Jayaraman, P.~K., and
  Furukawa, Y.
\newblock {SkexGen}: Autoregressive generation of {CAD} construction sequences
  with disentangled codebooks.
\newblock In \emph{International Conference on Machine Learning (ICML)}, 2022.

\end{thebibliography}
\bibliographystyle{icml2024}

\newpage
\appendix
\onecolumn

\begin{figure*}[t!]
\centering
  \includegraphics[width=.8\textwidth]{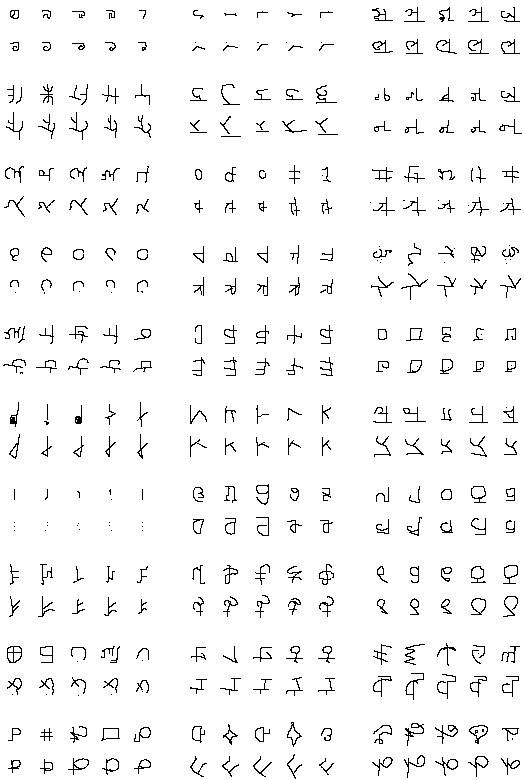}
  \caption{
    Qualitative examples of unconditional concept generations on the Omniglot domain.
    We show 30 concepts synthesized by our method where each concept is associated with two rows of five images. The bottom five images depict five samples from each concept, and the top five images show the nearest neighbor in the training set by Chamfer distance to each sample. 
    } 
  \label{fig:uncond_con_gen}
\end{figure*}

\section{Appendix Overview}

We overview the contents of our appendices.
In Section~\ref{sec:app_more_results} we provide additional experimental results.
In Section~\ref{sec:app_domains} we provide more information concerning our various visual domains.
In Section~\ref{sec:app_net_details} we provide details of our learned models.
In Section~\ref{sec:app_train_details} we provide details on how we design our training procedure.
In Section~\ref{sec:app_exp_details} we provide further details of our experimental design.
Finally, in Section~\ref{sec:app_baseline} we describe implementation considerations of each alternative we compare our system with.

\section{Additional Results}
\label{sec:app_more_results}

\subsection{Out-of-distribution Few-shot Generation}
\label{sec:app_layout_gen}

\begin{figure*}[t!]
    \centering
    \tiny
    \setlength{\tabcolsep}{.5pt}
    \begin{tabular}{cccccccccccccccccc}

        \raisebox{1.0em}{NN Train} \hspace{.25em} &
        \includegraphics[{width=.058\linewidth}]{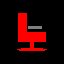} &
        \includegraphics[{width=.058\linewidth}]{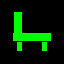} &
        \includegraphics[{width=.058\linewidth}]{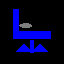} &
        \includegraphics[{width=.058\linewidth}]{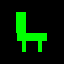} &
        \includegraphics[{width=.058\linewidth}]{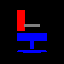} &
        \hspace{.5em}
        &
        \includegraphics[{width=.058\linewidth}]{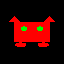} &
        \includegraphics[{width=.058\linewidth}]{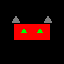} &
        \includegraphics[{width=.058\linewidth}]{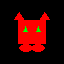} &
        \includegraphics[{width=.058\linewidth}]{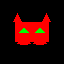} &
        \includegraphics[{width=.058\linewidth}]{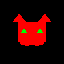} &
         \hspace{.5em}
        &
        \includegraphics[{width=.058\linewidth}]{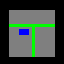} &
        \includegraphics[{width=.058\linewidth}]{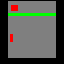} &
        \includegraphics[{width=.058\linewidth}]{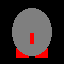} &
        \includegraphics[{width=.058\linewidth}]{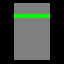} &
        \includegraphics[{width=.058\linewidth}]{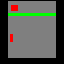} \\

        \raisebox{1.0em}{Input} \hspace{.5em} &
        \includegraphics[{width=.058\linewidth}]{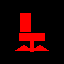} &
        \includegraphics[{width=.058\linewidth}]{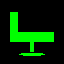} &
        \includegraphics[{width=.058\linewidth}]{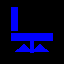} &
        \includegraphics[{width=.058\linewidth}]{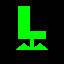} &
        \includegraphics[{width=.058\linewidth}]{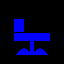} &
        \hspace{.5em}
        &
        \includegraphics[{width=.058\linewidth}]{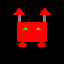} &
        \includegraphics[{width=.058\linewidth}]{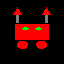} &
        \includegraphics[{width=.058\linewidth}]{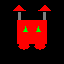} &
        \includegraphics[{width=.058\linewidth}]{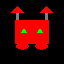} &
        \includegraphics[{width=.058\linewidth}]{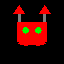} &
         \hspace{.5em}
        &
        \includegraphics[{width=.058\linewidth}]{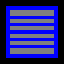} &
        \includegraphics[{width=.058\linewidth}]{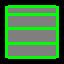} &
        \includegraphics[{width=.058\linewidth}]{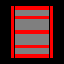} &
        \includegraphics[{width=.058\linewidth}]{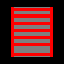} &
        \includegraphics[{width=.058\linewidth}]{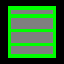} \\
        \raisebox{1.0em}{arVHE} \hspace{.5em} &
        \includegraphics[{width=.058\linewidth}]{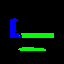} &
        \includegraphics[{width=.058\linewidth}]{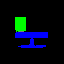} &
        \includegraphics[{width=.058\linewidth}]{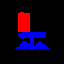} &
        \includegraphics[{width=.058\linewidth}]{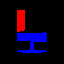} &
        \includegraphics[{width=.058\linewidth}]{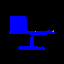} &
        \hspace{.5em}
        &
        \includegraphics[{width=.058\linewidth}]{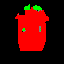} &
        \includegraphics[{width=.058\linewidth}]{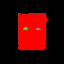} &
        \includegraphics[{width=.058\linewidth}]{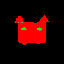} &
        \includegraphics[{width=.058\linewidth}]{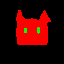} &
        \includegraphics[{width=.058\linewidth}]{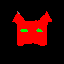} &
         \hspace{.5em}
        &
        \includegraphics[{width=.058\linewidth}]{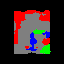} &
        \includegraphics[{width=.058\linewidth}]{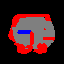} &
        \includegraphics[{width=.058\linewidth}]{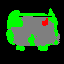} &
        \includegraphics[{width=.058\linewidth}]{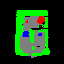} &
        \includegraphics[{width=.058\linewidth}]{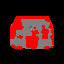} \\
        \raisebox{1.0em}{Ours} \hspace{.5em} &
        \includegraphics[{width=.058\linewidth}]{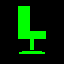} &
        \includegraphics[{width=.058\linewidth}]{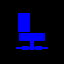} &
        \includegraphics[{width=.058\linewidth}]{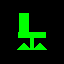} &
        \includegraphics[{width=.058\linewidth}]{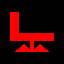} &
        \includegraphics[{width=.058\linewidth}]{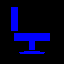} &
        \hspace{.5em}
        &
        \includegraphics[{width=.058\linewidth}]{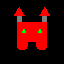} &
        \includegraphics[{width=.058\linewidth}]{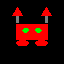} &
        \includegraphics[{width=.058\linewidth}]{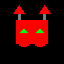} &
        \includegraphics[{width=.058\linewidth}]{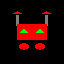} &
        \includegraphics[{width=.058\linewidth}]{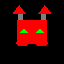} &
         \hspace{.5em}
        &
        \includegraphics[{width=.058\linewidth}]{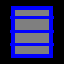} &
        \includegraphics[{width=.058\linewidth}]{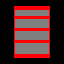} &
        \includegraphics[{width=.058\linewidth}]{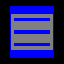} &
        \includegraphics[{width=.058\linewidth}]{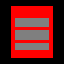} &
        \includegraphics[{width=.058\linewidth}]{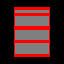} \\
        &\multicolumn{5}{c}{\textit{Combination Generalization (easy)}} & \multicolumn{6}{c}{\textit{Attribute Generalization (medium)}} & \multicolumn{6}{c}{\textit{Category Generalization (hard)}}  \\
    \end{tabular}
    \caption{Qualitative few-shot generation results that demonstrate our method's ability to generalize to out-of-distribution concepts, see Section~\ref{sec:app_layout_gen}.} 
    \label{fig:qual_lay}
\end{figure*}

As discussed in Section~\ref{sec:res_discussion}, we designed the Layout domain so that we could evaluate the out-of-distribution generalization capabilities of different approaches.
We visualize few-shot generations that different methods make for the layout domain for concepts that gradually get more and more out-of-distribution in Figure~\ref{fig:qual_lay}.
From left-to-right, we present example few-shot generations for an easy, medium and hard concept.
The easy concept (a side facing chair) has a set of attributes that have all individually been seen in the training set, but presents them in a new combination.
The medium concept (a crab) introduces a new attribute not seen in the training set: extended and vertical arms.
The hard concept (a bookshelf) introduces a new meta-concept that was never seen in the training set.
The second row of the figure show the input prompt set, where in the top row we show the nearest neighbor in the training set to each image in the prompt, according to our reconstruction metric.
On the third row we show generations produced by the arVHE comparison condition, while on the bottom row we show generations produced by our method.
While arVHE does reasonably well on the easy case, as the input prompts get more and more out-of-distribution it begins to generate nonsensical outputs.
On the other hand, our approach scales much better to out-of-distribution inputs, even though they don't match any images from the training set.

\subsection{Method Ablation Study}
\label{sec:app_ablation}

We run an ablation study to validate different design decisions of our method. We compare our described system against the following variants.
\textit{Ours - rel} is a variant of our method where we remove parameter relationships from Template Programs.
As by default we only support parameter relationships for argument types that take on discrete values (i.e. categorical variables) we also investigate a variant of our system that adds parameter relations (static assignment and reuse) for float-typed arguments: \textit{Ours + float rels}. 
We also compare against a version of our method where we remove~\HOLE~tokens, so that instantiations from Template Programs always use the same function call sequence: \textit{Ours - \HOLE}.
Here, we task our network to specify a single program structure that is applicable across the group without using~\HOLE~tokens, and it is still responsible for declaring parameter relationships. As there are no~\HOLE~tokens, the ExpansionNet will not be used, but the ParamNet will still be used to figure out how the instantiations of the Template Program should be parameterized. 
Next we compare against a variant where we remove the Structural Expansion step, so the ParamNet must produce a program from the Template Program directly.
As it doesn't see the \Expansion~intermediary result, it must fill in~\HOLE~tokens while figuring out how to predict parameter values.
We call this variant \textit{Ours - \Expansion}.
Finally, we compare against a variant of our base method without any finetuning, where networks only get to train on synthetic data: \textit{Ours - finetune}

We evaluate these ablation conditions on the 2D layout domain, and report results of our experiments in Table~\ref{tab:abl_table}. 
We compare our method against these variants with respect to few-shot generation performance (\textit{FD, MMD, Cov}), co-segmentation performance (\textit{mIoU}), and how well the inferred results optimize our objective~\Objective.
For ease of interpretation, we report all results as a percentage of the performance reached with respect to our default version (100\%).

As shown, our default method achieves the best performance along all of these tracked metrics.
The variant without finetuning clearly does the worst, as these networks are not specialized for the target dataset.
The results of this experiment validate our parameter relations design: keeping relations for discrete-valued parameters outperforms either no parameter relations or adding relations for float-valued parameters.
Using the~\HOLE~construct improves performance quantitatively. Moreover this construct is needed to capture complex input concepts that have more than a single expression mode.
For instance~\HOLE~tokens are required to model the chair concept with armrests and either a regular or pedestal base shown in the bottom left of Figure~\ref{fig:qual_main}. 
Finally, this ablation experiment demonstrates that our decision to use Structural Expansions simplifies the task of the ParamNet; we hypothesize this result is due to the fact that when attending over a~\Expansion, in contrast to attending over the~\Template, all of the functions and parameter-types that will be used in the end instantiation are known.

\begin{table*}[t!]
    \centering    
    \caption{Comparing ablated versions of our method to our default settings. Each metric is reported as a percentage, with respect to the performance our default approach achieves. See Section~\ref{sec:app_ablation} for details.}
    \vspace{1em}
    \begin{tabular}{@{}lccccccc@{}}
		\textbf{Method} & \textbf{FD} & \textbf{MMD} & \textbf{Cov} &  & \textbf{mIoU} & & \Objective \\
		\midrule
		  Ours & \textbf{100\%} & \textbf{100\%} & \textbf{100\%} && \textbf{100\%} && \textbf{100\%} \\
        Ours - rels	&	78.5\% &	92.6\% &	96.8\% &	&89.3\% &	& 95.8\% \\
        Ours + float rels &	93.9\% &	96.7\% &	98.2\% & &	96.7\% & &	95.3\% \\
        Ours - \HOLE &	96.3\% &	98.3\% &	97.6\% &	& 97.4\% & &	98.0\% \\
        Ours - \Expansion & 80.3\% &	89.9\% &	94.3\% & &	86.5\% & &	94.7\% \\
        Ours - finetune	 &	57.6\% &	80.5\% &	35.0\% & &	70.7\% &	& 81.6\%  \\ 
        \bottomrule
    \end{tabular}
    \label{tab:abl_table}
\end{table*}

\subsection{Unconditional Concept Generation}
\label{sec:app_uncond_con_gen}

As we mention in Section~\ref{sec:res_discussion} our Template Program framework is able to sample novel concepts unconditionally.
We visualize some concepts that our method is capable of producing in Figure~\ref{fig:uncond_con_gen}.

To produce these visualizations, we use the networks trained during the wake-sleep phase of our finetuning process,~\gennets.
Using the version of our TemplateNet from~\gennets~that does not condition on visual information, we first sample a Template Program.
Then using the ExpansionNet and ParamNet from~\gennets~that condition only on program inputs, we sample five program instantiations from this Template Program. 
Each bottom row in the figure shows the executed versions of these five samples, and above each sample we show the nearest neighbor character in the training set according to our reconstruction metric. 

\subsection{Visual Concept Groupings}

Typically, past concept learning approaches have assumed access to a dataset that is structured according to visual concepts. For instance, systems like VHE or FSDM require the ability to sample groups of input from the same visual concept during training. This is the same amount of dataset structure that our method requires: during fine-tuning we randomly sample “tasks” according to these visual concept groupings. Note that this requirement is less stringent than many inverse procedural modeling systems, and the BPL and GNS systems, that additionally require per-object structural annotations.

The Omniglot dataset was designed with this kind of visual concept decomposition in mind: each example data-point corresponds with exactly one character type. We design our layout domain in the spirit of Omniglot: each image in the layout domain is associated with a single concept. Following past work, on these domains we always assume “valid” input groups, such that each member is from the same visual concept.

However, this type of clean partition is not as easy to find for 3D shape structures. As there are no known datasets that group shape structures into visual concepts, we propose a heuristic method for forming approximate visual concepts out of shape structures (Appendix ~\ref{sec:app_domain_shape}). 
The concept groups we find under this formulation have different levels of consistency among their members (where we say a less consistent group forms a “harder” input problem).

For instance consider the examples shown in Figure~\ref{fig:qual_main}.
A chair with a regular base and vertical slats (row 7, col 6) could be in one group with only chairs that also have regular bases and vertical slats (row 7, col 7) or (like in the example we show) could also be grouped with chairs that have backs with horizontal slats (row 7, col 8).
In our paradigm, the group of visual inputs (along with our objective function) implicitly defines the granularity of the target visual concept. In this case, the latter grouping is considerably harder to handle for concept learning tasks, as it requires a method that is able to reason over input groups that partially mismatch on structures. 

Our Template Programs framework is capable of handling even difficult input groups; our partial program formulation allows our system to explicitly maintain the shared structural aspects of the group while leaving~\HOLE~tokens as responsible for representing the aspects of the input group that structurally differ. This design allows us to successfully capture the visual concepts of the two chair groups in Figure~\ref{fig:qual_main}. The left chair group has filled in chair backs, arm-rests, but alternates between regular and pedestal chair bases. The right chair group has regular leg bases, no arm-rests, but differs between chair backs with horizontal slats or vertical slats. As can be seen in the “gen” row, our system is capable of synthesizing novel shape structures that accord with the structural specifications implied by the input visual groups.

\subsection{Reconstruction Performance}

Our system learns how to amortize the difficult inverse search problem of finding a Template Program and instantiations that correspond with a group of visual inputs. 
This search (our inference procedure) is guided by our networks which are trained on a “training corpus” of visual concepts, separate from those we evaluate on.

The ``seg" rows in Figure~\ref{fig:qual_main} visualize the reconstructions (of the inputs on the top rows) that our method produces. 
While these reconstructions do not exactly recreate the input, they usually create very good approximations. 
If reconstruction was our primary goal, it might even be possible to improve the fit through a differentiable execution and refinement procedure.

To explore this phenomenon further, we provide the following reconstruction performance results across our domains in Table~\ref{tab:recon_table}. 
We report the reconstruction fit for both the training set and test set visual concepts. 
To show the benefits of our learning methodology we compare the reconstruction fit from the pretrained version of our networks (that learn only on synthetic data) to the finetuned versions of our networks (that finetune on visual concepts from training set).
The metrics we use are (full descriptions in Appendix~\ref{sec:app_domains}):
\begin{itemize}
    \item \textbf{2D Layout}: color-based IoU (higher is better)
    \item \textbf{Omniglot}: edge-based chamfer distance (lower is better)
    \item \textbf{3D Shapes (primitive input)}: structural corner distance (lower is better)
    \item \textbf{3D Shapes (voxel input)}: IoU (higher is better)
\end{itemize}

\begin{table*}[t!]
    \centering    
    \caption{Comparing reconstruction performance across domains, concept sets, and model versions.}
    \vspace{1em}
    \begin{tabular}{@{}lccccc@{}}
		\textbf{Domain} & \textbf{Mode} & \textbf{Train Recon} & \textbf{Test Recon} & \textbf{Test Recon (long)} \\
		\midrule
        \textit{2D Layout} $\uparrow$ & Pretrain & .822 & .808 & \\
         & Finetune & .972 & .909 & .937 \\
         \midrule
         \textit{Omniglot} $\downarrow$ & Pretrain & .658 & .648 & \\
         & Finetune & .468 & .503 & .405\\
         \midrule 
         \textit{3D Shapes (prim)} $\downarrow$ & Pretrain & .26 & .305 & \\
         & Finetune & .05 & .06 & .05 \\
         \midrule
         \textit{3D Shapes (voxel)} $\uparrow$ & Pretrain & .601 & .589 & \\
         & Finetune & .865 & .83 & .851 \\

        \bottomrule
    \end{tabular}
    \label{tab:recon_table}
\end{table*}

As demonstrated, our solution is effective at solving this inverse visual program induction problem. For both the training concepts and the held-out test concepts, our finetuning procedure meaningfully improves the reconstruction performance in all cases. For our downstream concept-related tasks we use a more expensive inference procedure (``long" - e.g. increase the beam size, Section~\ref{sec:met_inf}) and this gives even better reconstruction results for test-set concepts (see the numbers in the rightmost column of the table).

While our system offers strong reconstruction performance, it is likely that alternative methods could be used to infer single visual programs that better reconstruct an individual visual input. In contrast, our system learns to solve this visual program induction problem over a group of inputs by going through a shared structural intermediary (a Template Program), which allows us to perform concept-related tasks like few-shot generation and co-segmentation (which prior single instance VPI approaches are not suited for).

\subsection{Failure Modes}

\begin{figure}[t]
    \centering
    \setlength{\tabcolsep}{.5pt}
    \begin{tabular}{cccccccccccc}
        {\raisebox{.75em}{\rotatebox{90}{Input}}} \hspace{.1em} &
        \includegraphics[{width=.085\linewidth}]{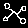} &
        \includegraphics[{width=.085\linewidth}]{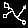} &
        \includegraphics[{width=.085\linewidth}]{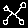} &
        \includegraphics[{width=.085\linewidth}]{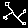} &
        \includegraphics[{width=.085\linewidth}]{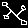} &
        \hspace{1em} &
        \includegraphics[{width=.085\linewidth}]{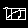} &
        \includegraphics[{width=.085\linewidth}]{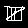} &
        \includegraphics[{width=.085\linewidth}]{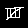} &
        \includegraphics[{width=.085\linewidth}]{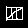} &
        \includegraphics[{width=.085\linewidth}]{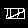} \\
        {\raisebox{.75em}{\rotatebox{90}{Recon}}} \hspace{.1em} &
        \includegraphics[{width=.085\linewidth}]{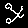} &
        \includegraphics[{width=.085\linewidth}]{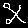} &
        \includegraphics[{width=.085\linewidth}]{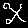} &
        \includegraphics[{width=.085\linewidth}]{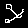} &
        \includegraphics[{width=.085\linewidth}]{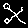} &
        \hspace{1em} &
        \includegraphics[{width=.085\linewidth}]{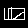} &
        \includegraphics[{width=.085\linewidth}]{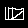} &
        \includegraphics[{width=.085\linewidth}]{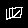} &
        \includegraphics[{width=.085\linewidth}]{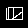} &
        \includegraphics[{width=.085\linewidth}]{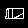} \\
        {\raisebox{.75em}{\rotatebox{90}{Gen}}} \hspace{.1em} &
        \includegraphics[{width=.085\linewidth}]{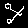} &
        \includegraphics[{width=.085\linewidth}]{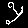} &
        \includegraphics[{width=.085\linewidth}]{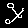} &
        \includegraphics[{width=.085\linewidth}]{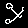} &
        \includegraphics[{width=.085\linewidth}]{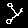} &
        \hspace{1em} &
        \includegraphics[{width=.085\linewidth}]{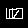} &
        \includegraphics[{width=.085\linewidth}]{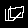} &
        \includegraphics[{width=.085\linewidth}]{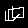} &
        \includegraphics[{width=.085\linewidth}]{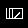} &
        \includegraphics[{width=.085\linewidth}]{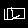} \\

    \end{tabular}
    \caption{When our method fails to find good reconstructions of an input concept, downstream task performance worsens.} 
    \label{fig:qual_recon_fail}
\end{figure}

\paragraph{Bad reconstruction}

A possible failure mode is that our inference networks can't find a Template Program whose instantiations well-capture an input visual group with respect to our objective function.
In such cases, the few-shot generation and co-segmentation results of our method are typically worse.
For instance, consider Figure~\ref{fig:qual_recon_fail}.  
For two Omniglot examples, in the top row we show the input concept groups, in the middle rows we show the reconstructions from our method, and in the bottom rows we show the few-shot generations from our method. 
Because the same Template Program is used in both the reconstruction and few-shot generation step, failure in one place often means failure in the other.
While from one perspective this is a limitation, a positive view of this phenomena is that our method can provide insight into cases where it is ``unsure" about its parse.
For instance, it could use the objective function score of its reconstruction as a measure of its confidence on how well it will perform on downstream tasks.
Moreover, as we show in Table~\ref{tab:recon_table}, reconstruction performance can improved by spending more time on inference, which can help to avoid this limitation.

\paragraph{Bad Input Groups}

How would our system handle `bad' input groups that contain outliers, or have no commonality among their members?
The job of the template network is to consume a group of visual inputs and infer a Template Program that captures the common structure among all members. In such an adversarial setting, it is possible (depending on random sampling) that there are no elements of structure common to all members of the input grouping.

In this case, the “best” result of our system would be to return a “dummy” Template Program that consists of a single~\HOLE~token; this~\HOLE~token would be able to be expanded into any arbitrary z to explain each individual group member. 

For typical visual concept groupings, this degenerate solution is discouraged by our objective function, which penalizes description length differences between "full" programs and their corresponding Template Programs. While finetuning our system with reasonable concept groups we have never observed the system falling-back to this degenerate solution.

Exploring how to extend our framework to handle “noisy” input groupings would be a very interesting direction for future work. This could potentially be approached by (i) extending our objective function to account for outliers (if we want to ignore the distractors) or (ii) adding control flow operators into the DSLs we learn over, which would give the Template Programs an opportunity to account for structural differences without relying solely on~\HOLE~tokens.

\section{Domain Details}
\label{sec:app_domains}

In this section we provide additional details on the visual domains we experiment on.
We describe the domain-specific languages our method uses and reconstruction metrics that guide our finetuning objective~\Objective.

For the 3D shape domain, we additionally provide details on how we produce our target dataset.
While we have previously explained how we divide concepts between training and test sets for each of our domains, we have not yet mentioned how we divide training examples into a validation set.
We find that a simple approach of taking a subset of training concepts with fixed exemplars as a `validation' set works well in practice. 
This validation set controls different early stopping components of our finetuning procedure, but otherwise these concepts are not given special treatment (i.e. they are not removed from the finetuning training set).

\subsection{Omniglot}

\paragraph{DSL}

We use the following domain-specific language for drawing Omniglot characters, where we present the notation with slight simplifications for ease of understanding.

\vspace{-2mm}
\begin{align*}
&START \xrightarrow{} GBlock;  \\
&GBLock \xrightarrow{} ONBlock \mid OFFBlock \mid MBlock \mid END \\
&ONBlock \xrightarrow{} \texttt{ON}; SBlock; GBlock \\
&OFFBlock \xrightarrow{} \texttt{OFF}; SBlock; GBlock \\
&MBlock \xrightarrow{} \texttt{MOVE}(si,mt,mf); GBlock \\
&SBlock \xrightarrow{} Stroke~\mid~\texttt{BOW}(bt,bf);~Stroke~\mid~\texttt{EMPTY}\\
&Stroke \xrightarrow{} \texttt{DRAW}(at,af,dt,df)\\
&si \in [0, 12] \\
&dt \in [0,8] / 8 \\
&at \in 360 * [0,8] / 8 \\
&bt \in 90 * [-2,2] \\
&mt \in [0,4] / 4 \\
&df \in [-2,2] / 40 \\
&af \in 9 * [-2,2] \\
&bf \in 30 * [-1,1] \\
&mf \in [-1,1] / 12 \\
\end{align*}

\vspace{-3mm}
The \texttt{ON} and \texttt{OFF} commands lift a pen on and off a virtual canvas; each series of strokes begins with one of these commands.
The \texttt{MOVE} command brings the pen back to a previous stroke, specified by a stroke index (\textit{si}) and a length along this stroke to travel specified by (\textit{mt, mf}). 
The \texttt{DRAW} command moves the pen at an angle specified by (\textit{at}, \textit{af}) for a distance of (\textit{dt}, \textit{df}). 
The trajectory of each \texttt{DRAW} command can be controlled by a \texttt{BOW} command which optionally pushes the trajectory inwards or outwards according to (\textit{bt}, \textit{bf}) parameter.
Even if making a curved stroke through the \texttt{BOW} operator, the end location of the pen is entirely controlled by the parameters of the \texttt{DRAW} command.

We draw attention to the fact that each real-valued parameter in this language is represented with a pair of arguments.
One member of each pair (those with \textit{t}) controls the coarse behavior, while the other member of the pair (those with \textit{f}) add a fine-grained delta to the initial coarse value (i.e. their values are combined through summation during execution).
This representation promotes consistency as close values will match on coarse binning token indices.
We further find it useful to treat these `coarse' real-valued parameters as categorical variables for the purposes of defining parameter relationships in the declaration of Template Programs, but we don't observe similar benefits when fine-grained values are included in this categorization
~(see ablation in Section~\ref{sec:app_ablation}). 
\HOLE~tokens are allowed to take place of any function. 

\textbf{Reconstruction Metric  } For our reconstruction metric~\Metric~, we use an edge-based Chamfer distance~\cite{sharma2018csgnet}. This allows us train our networks without access to stroke data, as we can compute this metric directly from binary images.

\textbf{Representational Capacity}
The maximum complexity of characters that our method is capable of representing is bounded by (i) the maximum number of tokens that our inference networks can handle and (ii) the maximum number of strokes we sample in the synthetic programs used in the pretraining step. This latter value is set to 12 in our sampling scheme, although through the introduction of HOLE tokens in Template Programs, some of the synthetic programs may end up using more than 12 stroke primitives. While the synthetic pretraining distribution will inform the behavior of the inference networks, this distribution will change over the course of bootstrapped fine-tuning and specialize towards ``real" Omniglot examples.

While we observe that these settings allow our model to reliably capture the majority of Omniglot characters, there are some very complex characters that might be hard to fit under these constraints with our top-down inference procedure. It should be possible to relax the constraints of both (i) and (ii), although the cost would be a larger GPU memory footprint and more complex pretraining data, which might require more training time and/or inference networks that use more parameters.

\subsection{2D Primitive Layout}

\paragraph{DSL}

We use the following grammar for creating layouts of 2D colored primitives. We present a slightly simplified representation of this language for clarity.

\vspace{-2mm}
\begin{align*}
&START \xrightarrow{} UBlock;  \\
&UBlock \xrightarrow{} \texttt{UNION}(ShBlock, UBlock) \mid ShBlock; \\
&ShBlock \xrightarrow{} (SymBlock \mid CBlock \mid MBlock \mid ScBlock); (PBlock \mid UBlock) \\
&SymBlock \xrightarrow{} \texttt{SymReflect}(axis) \mid \texttt{SymRotate}(n) \mid \texttt{SymTranslate}(n, xt, xf, xt, yf)   \\
&CBlock \xrightarrow{} \texttt{Color}(ctype) \\
&MBlock \xrightarrow{} \texttt{Move}(xt, xf, yt, tf) \\
&ScBlock \xrightarrow{} \texttt{Scale}(wt, wf, ht, hf) \\
&PBlock \xrightarrow{} \texttt{Prim}(ptype) \\
&axis \xrightarrow{} X\: \mid\: Y  \\
&ctype \xrightarrow{} red \mid green \mid blue  \\
&ptype \xrightarrow{} square \mid circle \mid triangle  \\
&n \in (1, 6)  \\
&xt \in [-3,3] / 4 \\
&yt \in [-3,3] / 4 \\
&wt \in .35 * [1,6] - .15 \\
&ht \in .35 * [1,6] - .15 \\
&xf \in [-2,3] / 20 - 0.025 \\
&yf \in [-2,3] /20 - 0.025 \\
&wf \in [-3,3] / 20\\
&hf \in [-3,3] / 20 \\
\end{align*}

Our language uses a \texttt{UNION} combinator to assemble a collection of primitives on a 2D canvas.
Primitives can take three types: squares, circles and triangles.
They are consumed by  \texttt{MOVE} and \texttt{SCALE} operators, where similar to our Omniglot domain, we make a distinction between the coarse and fine parts of each real-valued argument.
Once again, we distinguish the coarse values with $t$ endings and the fine values with $f$ endings.
Our motivations for adopting this tiered representation for real-values are identical to the Omniglot setting.
Instantiated primitives are colored grey, but can change color when passed through a \texttt{COLOR} operator. 
Our DSL also supports symmetry operations: \texttt{SymReflect} creates a reflectional symmetry group over a specific axis. \texttt{SymRotate} creates $n$ copies of its input argument about the origin. 
\texttt{SymTranslate} creates $n$ copies of its input argument in a direction that is parameterized by a distance in the same way as \texttt{MOVE}.

\textbf{Reconstruction Metric} For the layout domain we use a color-based intersection over union metric. Given two images, we first identify all of the occupied pixels, and which of our four colors each occupied pixel is filled in with. We then calculate the `intersection' numerator between these two images by counting the number of pixels that are both occupied with the same color.
We calculate the `union' denominator between these two images by counting any pixel in either image that is occupied. Our final value~\Metric~is calculated by dividing the numerator by the denominator.
\HOLE~tokens are allowed to take the place of any function.

\subsection{3D Shape Structures}
\label{sec:app_domain_shape}
\paragraph{DSL}

We use the following domain-specific language for 3D shape structures, which is adapted from~\citet{jones2020shapeAssembly}. We present a slightly simplified representation of this language for clarity.

\vspace{-2mm}
\begin{align*}
&START \xrightarrow{}  BBoxBlock; ShapeBlock; \\
&BBoxBlock \xrightarrow{} \text{bbox} = \texttt{Cuboid}(x, x, x)  \\
& ShapeBlock \xrightarrow{} (PBlock ; ShapeBlock) \mid \texttt{FILL}~\mid~\texttt{END}  \\
&PBlock \xrightarrow{}  CBlock; Attach; SBlock \\
&CBlock \xrightarrow{} c_n = \texttt{Cuboid}(x, x, x)~\mid~c_n = START \\  
&Attach \xrightarrow{} \texttt{attach}(cube_{n}, f, uv, uv) \\
&SBlock \xrightarrow{} Reflect \mid Translate \mid None \\
&Reflect \xrightarrow{} \texttt{Reflect}(\text{axis}) \\
&Translate \xrightarrow{} \texttt{Translate}(\text{axis}, m, x) \\
&f \xrightarrow{} right\: \mid\: left \:\mid\: top\: \mid\: bot\: \mid\: front\: \mid\: back \\
&\text{axis} \xrightarrow{} X\: \mid\: Y \:\mid\: Z\: \\
& x \in [0, 40] / 40. \\
& uv \in [0, 20]^2 /20. \\
& n \in [0, 4] \\
& m \in [1, 5] \\
\end{align*}

This DSL creates shape structures by defining cuboids, and arranging them through attachment.
Cuboids are instantiated with the \texttt{Cuboid} command.
Each \texttt{Attach} command moves one command to connect to previous part, indicated by $cube_{n}$ at a location specified by the other parameters of the command.
This language supports the creation of reflectional symmetry groups (\texttt{Reflect}) and translational symmetry groups \texttt{Translate}. 
Of note, we allow the DSL to expand hierarchically, so that Cuboids can become the bounding volume of their own sub-program (represented above with the return to the \textit{START} block).
These nested sub-programs are allowed to be set to a completely filled mode (\texttt{FILL}) or instead expand into empty space if immediately followed by the \texttt{END} operator.
Differing from other languages, we only allow~\HOLE~tokens to replace these \textit{START} tokens that define sub-program structures, to better match the hierarchical processes by which manufactured shapes are commonly modeled.

\paragraph{Recon Metric} We employ different metrics for this domain dependant on the visual representation.
When we operate over 3D voxel fields, we simply use the voxel occupancy intersection over union as our metric~\Metric.
When we operate over primitive soups, i.e. unordered collections of primitives, we use the following matching procedure: we first calculate the pairwise distance between each primitive by calculating the bidirectional Chamfer distance on the sets of corner points that form each cuboid. 
Assuming the two shapes we are comparing contain N and M cuboids, we converted these distances into a NxM array, and find an optimal matching through the Hungarian matching algorithm~\cite{kuhn1955hungarian}. 
Our metric~\Metric~is then calculated as the mean value of the entries of the matrix that form this assignment. 
When N $!=$ M, we convert the distance array into a square matrix using the larger dimension, filling in the `non-matched' entries with a high default value that penalizes structural mismatch.

\paragraph{Target Data} We source input shape structures by leveraging the structural annotations provided in the PartNet dataset~\cite{PartNet}. 
As our DSL supports only axis-aligned parts, we filter out any shape structures that require other kinds of oriented cuboids.
We then make use of the parsing procedure introduced by~\citet{jones2020shapeAssembly} to heuristically find ShapeAssembly programs, under the original DSL formulation, that correspond with these input shapes.
We try converting these programs into our DSL formulation, and check the geometric similarity between this execution and the original PartNet shape, as a sanity check to see if this shape structure \textit{could} be modeled under our procedural language.

At the end of this preprocessing stage, we are left with over 10,000 shapes from the chair, table and storage classes of PartNet.
We use the corresponding parsed ShapeAssembly programs to group these shapes into concept groups.
We differentiate the internal group consistency along 2 axes: whether or not the group would likely require a~\HOLE~token and whether or not the group would have a consistent application of attachment commands. 
We parse concept groups under all four combinations of these difficulty settings, choosing 25 concept groups from each setting to populate our test set, where each concept is `formed' according to a grouping of 10 exemplars.
We treat all other shapes not assigned to the test set as training shapes, and during finetuning we randomly sample concept groupings from this set according to the same concept identification procedure.

\section{Model Details}
~\label{sec:app_net_details}

\subsection{Architecture Details}
All of our auto-regressive networks are implemented as standard Transformer decoder-only models~\cite{att_is_all}. 
We use learned positional encodings, these cap the maximum sequence lengths for the various networks. 
There are three positional encodings for various sequences: the Template Program sequence, the Structural Expansion sequences, and parameter instantiation sequences.
For the layout domain we cap these at sizes: (64, 16, 72), for the omniglot domain we cap these at sizes (64, 16, 64), for the shape domain we cap these at sizes of (64, 24, 80).

\textbf{Visual Encoders}
We employ encoder networks that convert visual inputs into latent codes, see Figure~\ref{fig:method}.

For the layout domain we use a standard CNN that consumes images of size 64x64x3.
It has four layers of convolution, ReLU, max-pooling, and dropout. Each convolution layer uses kernel size of 3, stride of 1, padding of 1, with channels (32, 64, 128, 256). 
The output of the CNN is a (4x4x256) dimensional vector, which we transform into a (16 x 256) vector.
This vector is then sent through a 3-layer MLP with ReLU and dropout to produce a final (16 x 256) vector that acts as an 16 token encoding of the visual input. 
The omniglot CNN is identical, except it uses one fewer convolution layer, a padding size of 2 in the final convolution layer, and its 3-layer MLP consumes features of size (16x128) and transforms them into size (16x256). 
In this way for Omniglot we also convert each input image into 16 visual tokens.

For the shape domain we have two different encoders depending on the input modality. 
For our 3D voxel model we follow a similar convolutional paradigm, extending all 2D convolutions to be 3D, changing the kernel size to 4, using padding of size 2, and adding an extra fifth convolution layer. 
When consuming voxel grids of size 64x64x64 this produces outputs of size (2x2x2x256), we send this through a 3-layer MLP to produce a (8x128) feature, that we reformat to be (4x256) in dimension. In this way, 3D shapes are represented with four visual tokens.

When we consume a primitive soup of input, we use a different architecture based on a Transformer encoder~\cite{att_is_all}.
We assume that each primitive is a cuboid with 6 dimensions that describe its 3D position and size.
We linearize these primitive attributes, and lift each of them to dimension 16 with a 2-layer MLP.
Following this we add a learned positional encoding to each attribute based on its attribute type.
We then have another `positional encoding' that is produced by concatenating all of the attributes of each primitive (in the lifted dimension) and sending this feature through a 2-layer MLP that outputs an embedding of the same size as the lifted dimension, which then gets summed back into each attribute. 
This scheme allows us to avoid worrying about how the primitives are ordered, while still allowing the attention scheme of the network to differentiate which attributes belong to which primitives.
We send this tokenized representation through a standard Transformer encoder network, where we prepend the sequence with four `dummy' tokens. 
Each token attends to every other token, and we treat the representations output in the indices of the four `dummy' tokens as the visual tokenization. 
These dummy tokens build up a representation that attends of the entire input in much the same way as [\textit{CLS}] tokens have been employed.
Note that this encoder assumes a maximum number of primitives as input, which we set to 20. If the input scene does not have 20 primitives, we leave these entries as zeros, and then don't attend over those corresponding positions in the sequence while encoding.

\subsection{Location Encoding scheme}
\label{sec:app_loc_scheme}
We adopt the location encoding scheme from~\citet{t5paper} for predicting how to file in \HOLE~tokens, while predicting each~\Expansion~, and parameter values, while predicting the complete~\program. 
Specifically, we use their notion of `sentinel' tokens to identify any locations in the linearized function sequence that need to be filled in autoregressively.
Then during each autoregressive step, we `prompt' the network to predict for a specific location by repeating the sentinel token. 
We depict examples of this process in Figure~\ref{fig:method}.
We treat each sentinel token as an independent token in our language, this limits the number of~\HOLE~and parameter tokens we can predict. 
We set the max number of~\HOLE~location encoding tokens to be 5, and the max number of parameter location encoding tokens to be 64.
Assigning a reuse parameter relationship in the~\Template~also uses similiar location encoding tokens: we allow for up to 4 of these \textit{shared tokens}: when multiple instances of any of these shared tokens appear in the~\Template~, we constrain instantiations of the~\Template~to assign these slots with matching parameter values.

\subsection{Generative Networks}
\label{sec:app_gcond_net}

\paragraph{Unconditional Generative Networks} 
We use unconditional generative networks to produce paired data during our wake-sleep step of fine-tuning. 
Specifically these networks are unconditional with respect to visual inputs, but they still condition on programmatic elements. 
These networks can also be used for unconditional concept generation, see Section~\ref{sec:app_uncond_con_gen}.
The networks we use for this process have an identical architecture to our inference networks. 
In fact, at the beginning of our fine-tuning process we initialize the weights of these networks with the weights of the inference networks that have undergone supervised pretraining. 
They differ from the inference networks by simply masking out (i.e. setting to 0) all of the visual latent codes that are used to condition the generation of the Template Program, the Structural Expansion and the final program. 
In this way, these networks only condition on token sequences, or in the case of the TemplateNet, don't attend over any prefix conditioning information.
Our training scheme for these networks uses the same losses as our training scheme for the inference networks, assuming we have paired data

\paragraph{Few-Shot Generative Networks} 
For few-shot generative tasks, we want a network that has conditioning information in between our inference networks (that condition on latent codes specific to visual inputs in an input group) and our unconditional generative networks (that don't condition on visual inputs).
To address this point, we train variants of our inference networks that condition on a mean-pooled latent encoding (i.e. we average the 5 visual latent codes that come from an input group). 
Note that this only affects the ExpansionNet and the ParamNet, as the TemplateNet already is designed to attend over an input visual group.
Once we create this mean-pooled latent encoding, the training procedure is undergone in the same fashion, except the shared latent code is used as conditioning information for all of the instances of the (\TemplateGroup,\ProgramGroup) pair. 
In this way, we task the network with learning to solve a one-to-many modeling problem: from the same conditioning information, the network has multiple valid targets.

This network is trained on the same paired data as our inference networks (the batches of data created by our ST, LEST and WS procedures). 
While its possible to train this network during finetuning alongside the inference network, we instead cache all of the training data our inference network consumes during finetuning, and then train this few-shot generative network in a separate process after our inference model has converged.
All of the few-shot generative results we demonstrate are sampled from these networks (after a Template Program describing an input group has been inferred).

\section{Training Details}
\label{sec:app_train_details}

We implement our networks in PyTorch~\cite{paszke2017automatic}.
We run all experiments on a NVIDIA GeForce RTX 3090 with 24GB of GPU memory, and 64 GB of RAM. 
During pretraining we set the batch size to max out GPU memory, this amounts to sizes of 32 for the 2D layout domain, 40 for Omniglot domain, 32 for the shape domain with a primitive soup input and 16 for the shape domain with voxel inputs (of size $64^3$). 
Note that this batch size is effectively multiplied by 5 for the ExpansionNet and ParamNet as we train on visual input groups of size 5.
During fine-tuning we set the batch size to 20 for all methods, except for the shape-voxels variant, which we set to 10 to avoid maxing out VRAM.

We use the Adam optimizer to train our networks~\cite{Kingma2014AdamAM} with a learning rate of 1e-4. 
We pretrain our networks on synthetic data sampled from each domain until we converge with respect to a validation set of similarly sampled synthetic paired data.
This takes approximately $\sim$~700k batches for the layout domain, $\sim$~600k batches for the shape domain, and $\sim$~300k batches for the Omniglot domain.

We finetune our inference networks with the procedure described in Section~\ref{sec:met_learn}.
For each concept in the training set, we sample a group of visual inputs (at random) from the concept, and record our inference results to produce the LEST and ST dataset. 
In this way if there are K concepts in~\Target, the size of the ST and LEST data on each training step will also be K. 
Differing from this, in the wake-sleep step of our finetuning procedure we can generate an arbitrarily large number of paired data by sampling our generative model.
We find that sampling a large number of `dreams' is helpful for our finetuning procedure, so we set the number of example~\Template~to sample in each training step to 30,000. This typically takes between 1 and 2 hours, differing slightly for each domain.
To encourage the `dreams' we sample to cover a wide-distribution, we design a negative rejection step where we resample any `dream' that either creates an already generated~\Template~or~\VisGroup. 
We find this rejection criteria is triggered at relatively infrequent rates ($\sim$5\% of the time).

Once we've created the ST, LEST and WS datasets, we use them to finetune our inference networks with cross entropy loss. 
We train over this datasets for multiple `epochs', where every 5th epoch we run our updated inference networks over concepts from the validation set.
We use the Objective~\Objective~from this validation inference to decide when to break out of the training step, and return to the inference step.
This early stopping inference procedure always backtracks to the version of the inference network that achieved the best~\Objective~measure on the validation set. We use a patience of 10 epochs, and finetune for at most 50 epochs.

Overall, we run our finetuning procedure to convergence for 25, 17, 32 inference-training loops for the layout, omniglot and shape domains respectively. 
This corresponded with 565, 450, 620 finetuning `epochs' for these domains. 
For the weights of our objective function~\Objective, we normalize each reconstruction metric to values typically between 0 and 1, and then we set $\lambda_1$ to 1.0 and $\lambda_2$ to 0.001. 
Moreover, when calculating the divergent description length between each Template Program and its respective program instantiations, we discard counting any parameter-types for which we don't support parameter relations. 
For instance, as we don't allow float variables to use parametric relations (see Section~\ref{sec:app_ablation}), we do not penalize these variables under~\Objective, because the~\Template~has no opportunity to constrain them.

\subsection{Token Sequence Formatting}
\label{sec:app_teacher_force}

Given a paired (\VisGroup,~\TemplateGroup,~\ProgramGroup) triplet we can produce training data for our inference networks.
We train under a teacher-forced autoregressive paradigm, where we make a single pass through the autoregressive network for each training batch.
The input for the TemplateNet is a linearized sequence of visual latent codes; these are randomized as we randomly order the visual inputs.
The target for the TemplateNet is the linearized sequence of tokens that describe the Template Program, where we use prefix notation to convert expression trees into flat sequences.
From~\Template~and~\program~pairs, we can derive targets for the ExpansionNet and the ParamNet. 
To find targets for the ExpansionNet, we simply identify mismatches in the functions that are used in the~\Template~versus the functions that are used in~\program: any expression tree in~\program~that is not found in the~\Template~must be the result of filling in a~\HOLE~token.
Similarly, we scan the~\Template~to identify any parameter relationships that have been defined, either in the form of specifying parameter arguments (static assignment) or using \textit{shared tokens}. 
As we know the final expression tree of the~\program~from its linearized form, we then use these declarative relationships to reformat the~\program~to replace all free parameters with \textit{sentinel tokens} (Section~\ref{sec:app_loc_scheme}).

\section{Experiment Details}
\label{sec:app_exp_details}

\subsection{Few-shot Generation}

\paragraph{Task design}
In the few-shot generation task we employ the following set-up.
For each concept in the test set of a particular domain, we take 5 examples from the concept, pass them as input into a method, and then ask the method to synthesize 5 new generations. 
We then compare these 5 generations to a separate set of 5 examples from same test-set concept (i.e. a reference set). 
As the layout domain is procedurally generated, we can sample more examples per concept, therefore in this domain we do the above procedure 5 times for each test set concept.
In this way for layout, our metrics compare sets of size 25 generations to 25 reference images (where these 25 generations came from 5 prompts).

\paragraph{Metrics}

We quantitatively evaluate few-shot generative capabilities (Table~\ref{tab:main_table}) with a series of metrics common to recent generative modeling approaches~\cite{achlioptas2018learning}.
Though these metrics are typically designed to operate over much larger sets, we think the trends they exhibit are indicative of few-shot generative performance (and their ordering is largely consistent internally).

Some of these networks directly compare the generated samples to a reference set for each concept. 
Frechet Distance (\textit{FD)}~\cite{FrechetInceptionDistance} measures the distributional similarity between two distributions of encodings.
Minimum Matching Distance (\textit{MMD}) measures the average minimum distance of each member of the reference set to any member of the generated set.
Coverage (\textit{Cov}) measures the percentage of reference set members who are the nearest neighbors to at least one member of the generated set.

We calculate all of the above metrics with respect to a latent space that is domain-specific.
To this end, for each domain, we train a visual auto-encoder to learn how to reconstruct `random' scenes from that domain.
For the layout domain these are randomly placed primitives.
For Omniglot, these are randomly placed strokes.
For shapes, these are randomly place cuboids primitives. 
We train each of these networks to convergence on 500,000 random scenes with a small bottleneck layer size (e.g. 100). 

For the layout and omniglot domain we train simple classifier networks to learn a K-way classification over all of the concepts present in the domain. 
For Omniglot, we train on 19 examples from each of the 1623 characters in the dataset, and hold out one example from each concept as a validation set.
Our classifier achieves a 82.4\% validation accuracy after convergence.
For layout, we train over 95 examples from each concept in a 20-way classification task over meta-concepts; we reach 99.9\% validation accuracy on a held out set of 5 examples per concept. 
The class confidence metric (\textit{Conf}) is then computed by taking each generated output, running it through the classifier, and then recording the probability that the classifier predicts for the index of the input concept.
Note that this metric is not dependant on the reference set of examples.

\subsubsection{Perceptual Study}
\label{sec:app_exp_percep_details}

\begin{figure}[t!]
\centering
  \includegraphics[width=.4\textwidth]{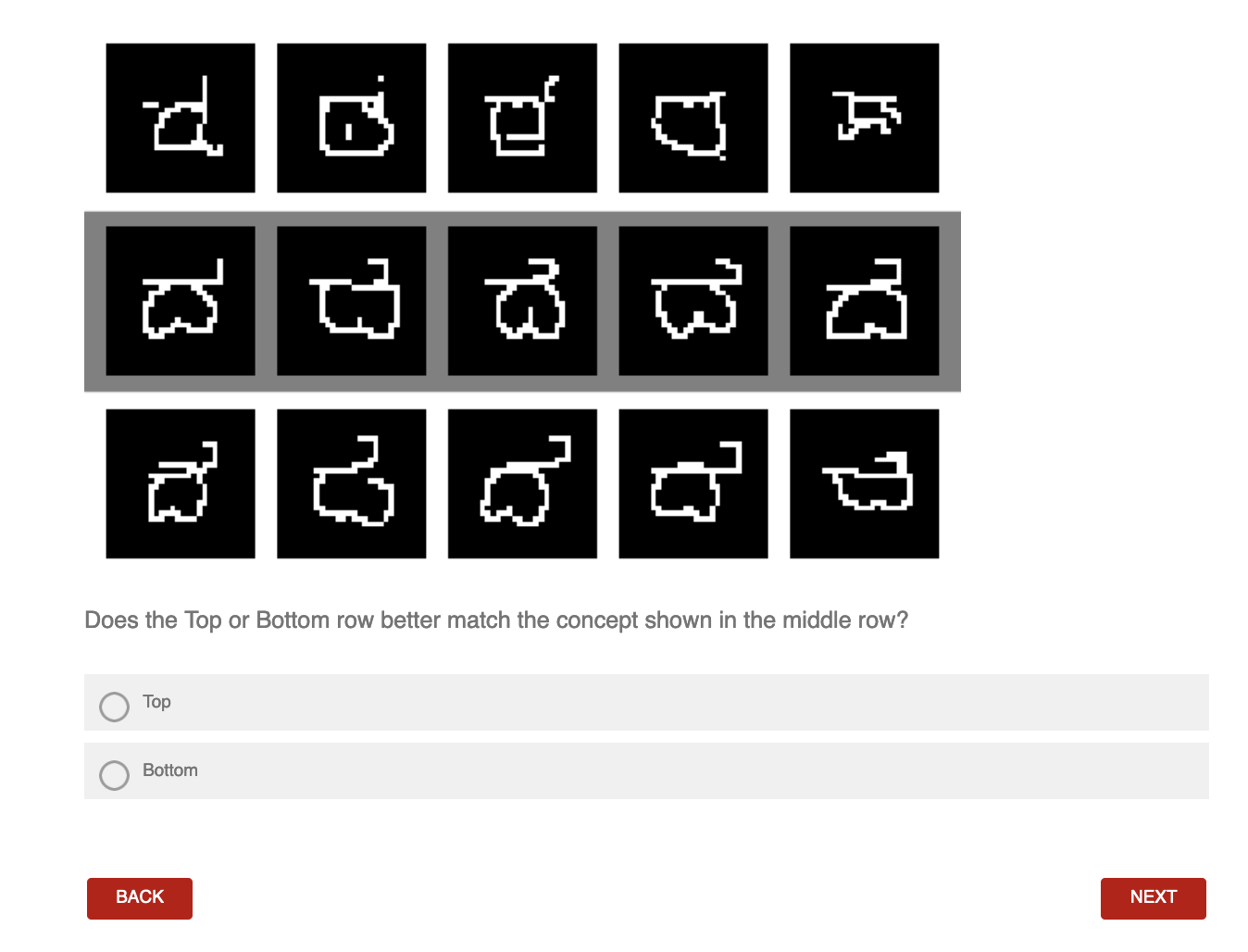}
  \caption{
    A visualization of the interface we use in our two-alternative forced-choice perceptual study.
    } 
  \label{fig:percep_study}
\end{figure}

We design a perceptual study to evaluate our method's few-shot generative capabilities.
Our study was designed as a two-alternative forced-choice questionnaire. 
We recruited 20 participants, who made decisions about which set of few-shot generations better matched a reference concept.

We show an example of our perceptual study interface in Figure~\ref{fig:percep_study}.
The middle row of each question shows the input prompt examples.
The bottom/top row are populated by the few-shot generations of competing methods based on the prompts shown in the middle row.
We randomize which method is shown on top vs bottom, and randomize the order of all examples within the row.

Participants were either shown 50 Omniglot character comparisons or 25 shape comparisons. We visualized shape comparisons with a simple rendering style of the primitive outputs produced by each method (for time considerations).

From our 20 participants we record 900 judgements of our method against three other conditions: ours vs arVHE for Omniglot (381 judgements), ours vs GNS for Omniglot (369 judgments) and ours vs arVHE for 3D shapes (150 judgements). We report the quantitative results from this study in Table~\ref{tab:percep}.

\subsection{Co-segmentation}
\label{sec:app_exp_coseg_details}

We formulate the co-segmentation task as follows.
We are given 5 examples as input, exactly one of these examples comes with a reference segmentation. 
The goal of each method is to propagate the labeling from this reference segmentation to the other members of the input group that lack a reference segmentation.
We show an example of this task in Figure~\ref{fig:qual_shape}.

We compare the produced segmentations against ground-truth annotations for each member of the input set. 
To quantitatively evaluate performance on this task we use a mean intersection over union metric (mIoU)~\cite{PartNet}.
This metric calculates the intersection over union for each label that appears in the ground-truth annotation, and then averages these values.

\subsubsection{Ground-Truth Segmentations}

Here we describe how we source ground-truth segmentations for each domain.

For 2D layouts, we produce these as a part of the way we design our meta-procedures. Each primitive group in these specifications is given a semantic label. We evaluate over all concepts in the test set.

For 3D shapes, we record the PartNet hierarchy annotation for each primitive of each shape structure we use~\cite{PartNet}.
Then within each test-set concept, we search for a group of 5 inputs that use the same semantic parts in their shape structures. 
If we find such a group, then this is the group from the concept we use during co-segmentation tasks.
From our 100 test set groups, we find such co-segmentation inputs for 94 of them.

We make use of Omniglot stroke data to produce the ground-truth segmentations for characters. 
We treat each stroke pattern broken by `BREAK' annotations as a separate segment~\cite{omniglot}.
Then, as humans vary in the ways that they order strokes to draw characters, for each test set character we run a clustering procedure to try to find valid and consistent segmentation groupings.
We first filter for finding groups of characters that use the same number of strokes, and more than a single stroke (otherwise the co-segmentation task is trivial).
Then we encode each stroke with a 4 dimensional feature: its length, its angle, its starting x position, and its starting y position.
We run an unsupervised clustering algorithm over this feature representation~\cite{dbscan}, identify if there is any cluster with more than 5 character members, and then take 5 characters from this cluster as a co-segmentation task (where our feature-wise distance creates a correspondence across the strokes of this group).
This automatic process generates 306 co-segmentation tasks from the 659 concepts in the Omniglot test set. 
We manually inspect the generated tasks, and filter out 22 cases where our clustering identified a group that did not have consistent stroke expression. 
This leaves us with 284 cosegmentation tasks that we use in our experiments.

\subsubsection{Group Parsing}

\paragraph{Template Programs} 
Template Programs support parsing by inferring instantiations from a shared~\Template~that explain a group of visual inputs.
As each instantiated program~\program~uses the function call structure specified by a Template Program, we can find correspondences in the visual outputs.
We create a corresponding group for each primitive type that the Template Program defines: these are created by the \texttt{PRIM} command for the layout domain, the \texttt{DRAW} command for Omniglot and the \texttt{Cuboid} command for 3D Shapes. Note that~\HOLE~tokens are always treated as a construct that creates primitive types. 
Any command that operates over this primitive type will inherent their corresponding part index (e.g. symmetry operations), excluding combinators like \texttt{Union}.

\paragraph{BAE-NET} BAE-NET creates corresponding group parses by performing an argmax over the last layer of an implicit network that is trained to solve occupancy tasks. This implicit network can be run over any spatial position, and assign this input point to one of its part `slots'.

\paragraph{BPL and GNS} The BPL and GNS methods perform one-shot parsing of input characters into an ordered collection of strokes.
This parsing is guided by their learned prior, which models how people produce characters.
Conscripting these methods to perform our co-segmentation task is a slight abuse of design, but as their output parses partition space in a consistent fashion, we think it a worthwhile comparison to make. 
Our method does not learn from any human demonstrations, so we are unable to solve the character parsing task as it is originally formulated~\cite{bpl,omniglot}.

\subsubsection{Label Propagation}

The parses we get from the above logic are consistent, but might not exactly recreate the input examples (if they do not achieve perfect reconstructions).
We thus employ a procedure, on a domain-by-domain basis, that propagates the parse from the reconstruction to its input example.
For the layout domain we first take the part index of each occupied pixel to match the primitive that last `covered it'. Then for any non-occupied pixels, we assign them to the closest instantiated primitive according to the distance from that pixel's center to the primitive.
For the shape domain, we take a similar approach, calculating the distance from voxel centers to each cuboid. 
For any voxel center that is occupied by more than one cuboid, we assign it to the occupying cuboid smallest in volume.
For Omniglot, we sample 200 points on each primitive stroke group. Then for five query points evenly spread out within each pixel location, we find the three closest points sampled from any stroke group. We tally up these votes for each pixel, and then each pixel is assigned to the primitive stroke group which recorded the most votes. Note that we employ this same procedure for our method, BPL, and GNS. 
BAE-NET doesn't need to employ this logic, as its parsing strategy operates over arbitrary input points by construction.

After we have this consistent parse for each region of the input group the procedure is almost done. 
We use the partitions from the labeled example to assign each parsed region a label.
Finally, we propogate this region-to-label mapping to all of the other examples in the input group.

\section{Comparison Method Details}
\label{sec:app_baseline}

We provide details on the methods we compare against.

\subsection{BPL~\cite{bpl}}

We use the author's released Matlab implementation: https://github.com/brendenlake/BPL. 
For five characters from each test-set concept we infer a parse, and use that parse to synthesize 1 new generation (in this way we create 5 few-shot generations from each group of 5). 
We wrapped this Matlab procedure with a python script, and ran it sequentially on a single machine, which took around 2 weeks.

\subsection{GNS~\cite{gns}}

We use the author's released implementation at https://github.com/rfeinman/pyBPL. 
We follow the same procedure as in BPL, inferring a parse for five characters from each test-set concept, and then using each parse to synthesize 1 new generation.

\subsection{FSDM~\cite{giannone2022fewshot}}

We follow the author's implementation, released at: https://github.com/georgosgeorgos/few-shot-diffusion-models. 
Unfortunately, the provided code was incomplete, and did not work out of the box. 
We made a best-effort attempt to fix these issues and run the model with the same procedure as described in the technical report.
We observed that this model was able to effectively produce few-shot generations for training characters, but struggled greatly on test-set concept generalization.

\subsection{VHE~\cite{hewitt2018variational}}

We attempted to use the author's implementation, released at: https://github.com/insperatum/vhe. 
Unfortunately the PixelCNN variant for Omniglot did not converge under training, we reached out to the authors, but they were unable to offer suggestions on how to fix these training issues.

Using the provided code as reference, we re-implemented the system with a simple CNN architecture, following the VAE framing as described in~\cite{sigmavae}.
Though we spent a fair amount of time tuning hyper-parameters, as evidenced by the quantitative results in Table~\ref{tab:main_table}, we were unable to achieve competitive performance.

\paragraph{arVHE} 

In an attempt to improve the performance of our VHE comparison condition, we implemented a related method that combines autoregressive models with the spirit of the VHE approach.
Specifically, we break down this few-shot generation modeling task into two separate stages. 
First we learn a domain-specific discretized representation.
For pixel and voxel input representations we use 2D and 3D CNNs in a vector-quantization scheme~\cite{vqvae}, so that we can convert each visual input into a sequence of discrete tokens.

We list the details of our VQ-VAE training:
for Omniglot we convert 28x28x1 images to a 7x7 grid of codes, under a dictionary of 64 codes with hidden dimension of 32.
For layout we convert 64x64x3 images to a 7x7 grid of codes, under a dictionary of 200 codes with hidden dimension of 100.
For shapes we convert 64x64x64 voxels to a 4x4x4 grid of codes, under a dictionary of 128 codes with hidden dimension of 64.
We try to use the smallest code-book size that can achieve near-perfect reconstructions for each domain.

Once we have trained this VQ-VAE for each domain, we can learn our arVHE model.
Like the VHE model, and our system, it learns by sampling random visual groups from the same concept. 
Following the procedure described in the VHE paper and code, we encode these visual concepts with a visual encoder, take a mean embedding, then use this embedding to condition an autoregressive generation process, where the goal is to predict a sequence of VQ-VAE tokens that correspond to another input example from the same concept.
We train this network with cross-entropy loss, on the discretized VQ-VAE tokens. 
For an apples-to-apples comparison against our method, the arVHE baseline uses the same visual encoders that our method uses (Section~\ref{sec:app_net_details}).
For predicting 3D shapes as a sequence of primitives, we instead just task the VQ-VAE model with predicting discretrized versions of each primitive attribute, where the primitives are randomly ordered (this allows us to skip the VQ-VAE step in this setting).

We note that this arVHE variant is a strong baseline method, outperforming VHE and FSDM in terms of quantitative metrics (Table~\ref{tab:main_table}).

\subsection{BAE-NET~\cite{chen2019bae_net}}

We follow the author's implementation released at: https://github.com/czq142857/BAE-NET. 
We take their architecture and training procedure and adapt it for each of our domains. 
BAE-NET has model implementations for 2D binary images and 3D voxel grids, so for these settings we directly use the method as described.
For the layout domain we have colored images that can adopt 4 color values (red, green, blue, or grey). 
In the default version of BAE-NET, it uses an MLP where the second to last layer is size NUM\_SEGS and the last layer is size 1; this 1 dimensional output learns a binary occupancy prediction for locations in space.
We modify the 2D BAE-NET version so that instead, the second to last layer is still size NUM\_SEGS, but the last layer is size 4; in this way we task BAE-NET to solve four binary occupancy problems at once, one for each of our colors. 
In the layout domain, we still take the part segmentation from BAE-NET by choosing the slot in the second to last layer that activates with the highest potential.

\end{document}